\documentclass[11pt,letterpaper]{article}
\usepackage[T1]{fontenc}
\usepackage[margin=1in]{geometry}
\usepackage{times}
\usepackage[authoryear,round]{natbib}
\setcitestyle{authoryear,round,citesep={;},aysep={,},yysep={;}}
\date{}

\usepackage{amsmath,amsfonts,bm}

\def\eqref#1{equation~\ref{#1}}

\def\1{\bm{1}}

\DeclareMathAlphabet{\mathsfit}{\encodingdefault}{\sfdefault}{m}{sl}
\SetMathAlphabet{\mathsfit}{bold}{\encodingdefault}{\sfdefault}{bx}{n}

\usepackage[normalem]{ulem}
\usepackage{wrapfig}
\usepackage{needspace}

\usepackage{amssymb}
\usepackage{booktabs}
\usepackage{array}
\usepackage{graphicx}

\usepackage{microtype}
\usepackage{xcolor}
\usepackage{xspace}

\usepackage[hidelinks]{hyperref}
\usepackage{url}

\newcommand{\tb}{\textsc{TempoBridge}\xspace}

\title{TempoBridge: Source-Conditioned Flow Matching with Optimal Transport Couplings for Single-Cell Population Transitions}

\author{%
\begin{minipage}{0.97\textwidth}
\centering
\normalsize
Bowen Han\textsuperscript{1}, Lingbei Meng\textsuperscript{2,3}, Shihuan Luo\textsuperscript{3},
Yupeng Zang\textsuperscript{4}, Wenlin LI\textsuperscript{2}\\[3pt]
Peize He\textsuperscript{5}, Yaodi Luo\textsuperscript{5}, Lian Zhang\textsuperscript{4},
Jianqing Zhu\textsuperscript{3}, Jinchao Xu\textsuperscript{2,3}\\[8pt]
{\small
\textsuperscript{1}Universiti Sains Malaysia\\[2pt]
\textsuperscript{2}The Chinese University of Hong Kong Shenzhen\\[2pt]
\textsuperscript{3}Shenzhen Loop Area Institute\\[2pt]
\textsuperscript{4}Shenzhen Research Institute of Big Data\\[2pt]
\textsuperscript{5}University of Electronic Science and Technology of China
}
\end{minipage}%
}
\hypersetup{pdfauthor={Bowen Han, Lingbei Meng, Shihuan Luo, Yupeng Zang, Wenlin LI, Peize He, Yaodi Luo, Lian Zhang, Jianqing Zhu, Jinchao Xu}}

\providecommand{\revised}[1]{#1}
\begin{document}

\maketitle
\begin{abstract}
Destructive single-cell measurements provide unpaired population snapshots rather than observations of the same cells across conditions. Local cell states and transition requests may also be insufficient to distinguish responses across source populations. We introduce TempoBridge, a common source-conditioned transport formulation for temporal, genetic, and chemical population transitions. Source cells initialize latent transport and provide a fixed empirical population summary. The velocity field receives this summary alongside the evolving cell state, flow time, and a structured transition descriptor. Minibatch optimal transport (OT) supplies couplings only for conditional flow-matching training paths; inference requires neither target expression nor OT computation. On held-out donors, TempoBridge achieves an Energy distance of 0.129 versus 0.144 for scGen. Genetic mean-expression $L_2$ error is 2.261 versus 3.156 for scGPT-scratch under Seen~2/2. On held-out compounds, condition-averaged drug-effect correlation is 0.598 versus 0.561 for the CellFlow adapter. 
Temporal ablations show higher mean distributional error after removing source context, replacing optimal transport with random pairing, or replacing flow matching with static residual regression. Together, these results demonstrate the predictive utility of a common source-conditioned transport formulation across held-out donors, gene combinations, and compounds.
\end{abstract}

\section{Introduction}
\label{sec:introduction}

Single-cell experiments measure populations at selected time points and under selected perturbations, leaving many differentiation states, gene combinations, and drug responses unobserved~\citep{cuomo2020singlecell,norman2019exploring,srivatsan2020massively}. We consider the prediction of a target gene-expression distribution from source-cell profiles and a requested time point or perturbation. Because sequencing destroys measured cells, training provides unpaired source and target populations rather than trajectories of the same cells~\citep{schiebinger2019optimal}. At inference, the requested target population is unavailable.

Methods for this setting use learned representations of cells and perturbations, mappings between populations, or conditional transport. scGen and CPA model perturbation effects in latent spaces~\citep{lotfollahi2019scgen,lotfollahi2023predicting}; CellOT maps unpaired populations~\citep{bunne2023learning}; and CMonge and CellFlow condition transport on experimental descriptors~\citep{driessen2026conditional,klein2025cellflow}. scDFM uses control-cell expression, while SCALE incorporates a set-level control representation~\citep{yu2026scdfm,chen2026scale}. These methods establish the value of source cells and experimental conditions. Here we ask how a transport field should use information about the particular source population as its cells move. Initializing a particle from a source cell gives the field that cell's state, but does not separately expose the population from which it was sampled.

The distinction becomes consequential when cells in similar states require different responses across source populations. Under a specified coupling and training-path construction, two populations can have different conditional mean velocities at the same latent state, flow time, and transition descriptor. A field conditioned only on these inputs receives the same input in both cases and cannot represent both velocities. We call this pointwise conflict \emph{source-dependent transition ambiguity} (Section~\ref{sec:source-ambiguity}). Supplying a source-population summary can resolve the conflict when that summary retains the relevant difference between populations.

We introduce TempoBridge, a transport model that keeps an empirical summary of the initial source population available throughout integration. Each source cell initializes a latent particle; mean pooling the initial cell embeddings produces a summary shared by all particles from that population. The velocity field takes the evolving particle state, flow time, source summary, and transition descriptor as separate inputs. We train it with optimal-transport conditional flow matching (OT--CFM), using minibatch Sinkhorn couplings to construct paths between unpaired endpoints~\citep{lipman2023flow,tong2024improving}. Prediction starts from source-cell latents and requires neither target expression nor OT computation. The temporal, genetic, and drug-response models follow this formulation but are trained independently.

We evaluate the complete models on held-out donors, gene combinations, and compounds. For intermediate populations of unseen donors, TempoBridge achieves an Energy distance of 0.129 versus 0.144 for scGen. On Seen~2/2 gene combinations, mean-expression $L_2$ error is 2.261 versus 3.156 for scGPT-scratch. On compounds excluded from training and validation, condition-averaged drug-effect correlation is 0.598 versus 0.561 for the CellFlow adapter. These comparisons measure the performance of the complete task models, but do not isolate the contribution of source context.

Targeted analyses separate source-context use from predictive benefit. In a controlled ambiguity task, context raises branch accuracy from 50.0\% to 98.0\%; donor-context swaps show that a fitted temporal model uses the summary. Matched temporal retraining lowers Energy from 0.139 to 0.135 and MMD from 0.397 to 0.378 with context, although effect metrics favor no context. Random pairing raises temporal Energy from 0.1291 to 0.1446; a static residual variant reaches 0.1350 but also changes the training objective. A separate drug conditioning analysis finds small gains in effect metrics but no consistent distributional benefit.

Our contributions are:
\begin{itemize}
    \item \textbf{A source-dependent ambiguity.} Under fixed couplings and paths, we identify a pointwise conflict for a velocity field given only cell state, flow time, and transition descriptor, and demonstrate it in a controlled construction.
    \item \textbf{A source-conditioned transport field.} A fixed summary of the initial source population conditions each particle's velocity throughout transport; OT couplings construct training paths without being required at inference.
    \item \textbf{An assessment of the design.} Donor-context swaps and matched context ablations distinguish use from metric-dependent benefit; temporal component variants test the training recipe, while independently trained models are evaluated on held-out donors, gene combinations, and compounds.
\end{itemize}

\section{Related Work}
\label{sec:related-work}
\label{app:related-work}

\paragraph{Optimal transport and flow matching.}
Optimal transport provides a distributional framework for learning from unpaired cell populations.
CellOT learns neural mappings between cellular distributions~\citep{bunne2023learning}, while CMonge conditions transport mappings on treatment covariates~\citep{driessen2026conditional}.
Flow matching learns continuous transformations through velocity regression~\citep{lipman2023flow}.
OT--CFM uses minibatch transport couplings to construct training paths~\citep{tong2024improving}, while GENOT uses flow matching to learn entropic Wasserstein and Gromov--Wasserstein transport plans~\citep{klein2024genot}.
For cellular response prediction, CellFlow incorporates experimental conditions into flow-based transport~\citep{klein2025cellflow}, and scDFM combines control-conditioned generation with MMD-based distributional alignment~\citep{yu2026scdfm}.
TempoBridge builds on OT--CFM and represents the evolving cell state, empirical source-population context, and structured transition descriptor as separate inputs to the velocity field.
Source cells initialize the latent particles, while a fixed summary of their initial population conditions the field throughout integration.

\paragraph{Temporal population modeling.}
Temporal single-cell models describe changes in population distributions from snapshots collected at different times.
Waddington-OT estimates temporal population couplings~\citep{schiebinger2019optimal}; PRESCIENT learns potential-based stochastic differentiation dynamics~\citep{yeo2021prescient}.
TrajectoryNet uses dynamic optimal transport~\citep{tong2020trajectorynet}, scNODE combines latent representations with neural differential equations~\citep{zhang2024scnode}, and BranchSBM models branching populations through multiple velocity fields and growth processes~\citep{tang2026branchsbm}.
TempoBridge uses endpoint supervision to predict intermediate population distributions at requested times for held-out donors.

\paragraph{Genetic and chemical perturbation prediction.}
Predicting responses to unseen interventions requires representations of the requested perturbations.
GEARS uses gene-relation graphs for multigene response prediction~\citep{roohani2024gears}; chemCPA encodes molecular information for novel-compound prediction~\citep{hetzel2022chemcpa}; and MAP combines knowledge-informed representations with a pretrained single-cell model~\citep{feng2026map}.
Genetic-perturbation benchmarks also motivate strong linear and additive baselines~\citep{ahlmanneltze2025baselines}.
TempoBridge represents genetic interventions as unordered gene sets and chemical interventions through molecular and experimental descriptors.

\section{Problem Setup}
\label{sec:problem-formulation}

\subsection{Unpaired Population Prediction}
\label{sec:conditional-transition}

For task $q\in\{\mathrm{time},\mathrm{gene},\mathrm{drug}\}$, a training record contains source cells $\mathcal X_s=\{x_i^s\}_{i=1}^{n_s}$, target cells $\mathcal X_t=\{x_j^t\}_{j=1}^{n_t}$, and a transition descriptor $c_q$.
Source and target populations are associated within each training record but have no cell-level correspondences. At inference, only $\mathcal X_s$ and $c_q$ are observed; the number of output cells is specified separately.
Each cell is represented by an expression vector in $\mathbb R^{G_q}$, where $G_q$ is the number of genes used for task $q$.
Their empirical distributions are
\begin{equation}
\widehat P_s=\frac{1}{n_s}\sum_{i=1}^{n_s}\delta_{x_i^s},
\qquad
\widehat P_t=\frac{1}{n_t}\sum_{j=1}^{n_t}\delta_{x_j^t},
\label{eq:empirical-distributions}
\end{equation}
where $\delta_x$ denotes a point mass at $x$.
Given the observed source distribution $\widehat P_s$ and a structured transition descriptor $c_q$, we aim to predict the corresponding target population distribution
\begin{equation}
P_{\theta_q}^{\mathrm{tgt}}
=
\mathcal T_{\theta_q}(\widehat P_s,c_q)
\approx
\widehat P_t.
\label{eq:conditional-population-prediction}
\end{equation}
The descriptor specifies the requested target; in temporal prediction, the field is conditioned on the shared Day~0--Day~3 endpoint transition, while the requested day determines the integration horizon rather than entering the field directly (Section~\ref{sec:tempobridge-condition-interface}).

\subsection{Source-Dependent Transition Ambiguity}
\label{sec:source-ambiguity}

Suppressing the task index, fix a source--target coupling and path construction for each training record, and let $Z_\tau$ and $U_\tau$ be the resulting latent position and path velocity.
For source population $\widehat P_s$ and descriptor $c$, define the conditional mean velocity $m_s(z,\tau,c)=\mathbb E[U_\tau\mid Z_\tau=z,\revised{\tau,}\widehat P_s,c]$.
If two source populations with the same $c$ have overlapping path states at $(z^\star,\tau^\star)$ but different values of $m_s(z^\star,\tau^\star,c)$, a deterministic field $v_\theta(z,\tau,c)$ receives identical inputs there and cannot match both means.
We call this pointwise conflict \emph{source-dependent transition ambiguity}.
\revised{For the fixed coupling and path law, Appendix~\ref{app:source-risk} quantifies the excess squared-error risk when source-specific mean velocities differ on an overlap of positive probability.}

Conditioning on an empirical summary of the source population gives
\begin{equation}
v_\theta\!\left(z,\tau,h(\widehat P_s),c\right).
\label{eq:source-conditioned-field}
\end{equation}
This field can distinguish the two population-specific mean velocities if $h$ retains the relevant source-population differences.
We test this information restriction in controlled ambiguity experiments (Section~\ref{sec:controlled-source-context}).

\section{TempoBridge}
\label{sec:tempobridge}

For each task $q$, a shared encoder maps unpaired source and target cells into a latent space. Source latents initialize transport, while their mean provides fixed source-population context for the velocity field throughout integration. Minibatch OT supplies latent endpoint pairs for flow matching. At inference, the learned field transports source-seeded particles under the requested  transition, and the decoder predicts expression without
target cells or OT.
Temporal, genetic, and drug-response models are trained separately.

\subsection{Cell-State Encoding and Source Context}
\label{sec:tempobridge-overview}
\label{sec:tempobridge-source-context}

\begin{figure}[!htbp]
\centering
\includegraphics[width=0.97\linewidth]{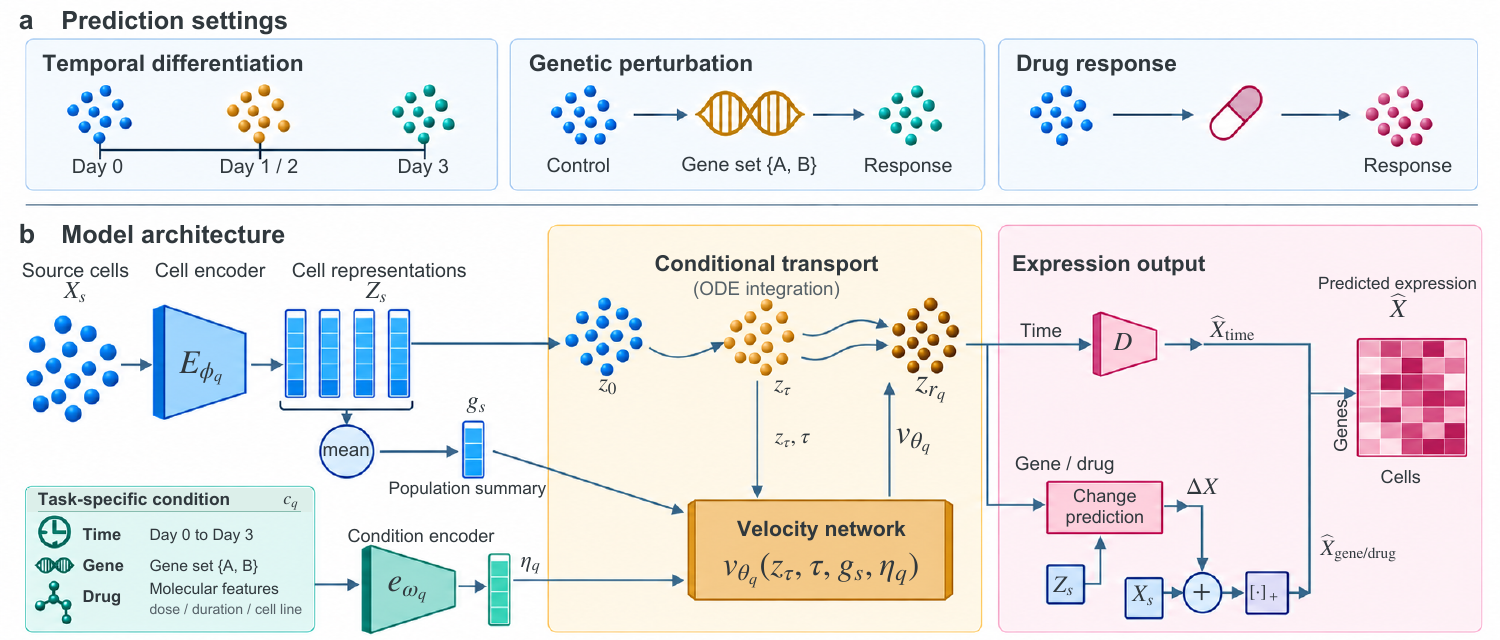}
\caption{
\textbf{TempoBridge overview.}
\textbf{(a)} Temporal, genetic, and drug-response models use a common source-conditioned transport formulation and are trained independently.
\textbf{(b)} Source cells initialize latent particles and provide a fixed empirical population summary.
The velocity field takes the evolving latent state, flow time, source summary, and encoded transition descriptor as inputs.
Each task model has its own expression decoder.
Minibatch OT constructs training paths; inference requires no target expression or OT computation.}
\label{fig:architecture}
\end{figure}

Figure~\ref{fig:architecture} summarizes the common transport design.
Within task $q$, an encoder
$E_{\phi_q}:\mathbb R^{G_q}\rightarrow\mathbb R^{d_q}$
maps source and target cells into the same latent space:
\begin{equation}
z_i^s=E_{\phi_q}(x_i^s),\qquad
z_j^t=E_{\phi_q}(x_j^t).
\label{eq:tempobridge-endpoint-encoding}
\end{equation}
Target cells are encoded during training; source latents seed transport (Section~\ref{sec:tempobridge-population-generation}).
The source population is summarized by permutation-invariant mean pooling~\citep{zaheer2017deepsets}:
\begin{equation}
g_s=h(\widehat P_s)=\frac{1}{n_s}\sum_{i=1}^{n_s}z_i^s.
\label{eq:tempobridge-source-context}
\end{equation}
Computed from the valid source cells, $g_s$ remains fixed  during integration and is shared by all particles from that population.

\subsection{Structured Condition Interface}
\label{sec:tempobridge-condition-interface}

The transition condition is encoded separately from the source population:
\begin{equation}
\eta_q=e_{\omega_q}(c_q^{\mathrm{field}}),\qquad
v_{\theta_q}(z_\tau,\tau,g_s,\eta_q),
\label{eq:tempobridge-condition-interface}
\end{equation}
where $c_q^{\mathrm{field}}$ is the descriptor supplied to the velocity field.
The field therefore receives the current cell state, flow time,
source-population summary, and transition descriptor as separate inputs.

For temporal differentiation, $c_{\mathrm{time}}^{\mathrm{field}}$
specifies the Day~0--Day~3 endpoint transition.
A request for day $t$ sets the integration horizon to
$r_{\mathrm{time}}(t)=\sqrt{t/3}$; this mapping is fixed rather than learned.
For genetic and drug prediction, $c_q^{\mathrm{field}}=c_q$ and the
integration horizon is one.
Genetic conditions represent one or two perturbation genes by
order-invariant pooling of entries from a task-specific gene-identity
embedding table.
Drug conditions use SMILES-derived molecular features~\citep{weininger1988smiles}, dose, treatment
duration, and cell line, without a learned drug-identity lookup.
Appendix~\ref{app:task-parameterization} gives the task-specific encodings.

\subsection{OT-Guided Conditional Flow Matching}
\label{sec:tempobridge-otcfm}

Source and target cells have no observed correspondence. Following flow matching and minibatch OT--CFM~\citep{lipman2023flow,tong2024improving}, we compute a coupling within each source--target population record. The cost is the squared Euclidean distance $C_{ij}=\lVert z_i^s-z_j^t\rVert_2^2$, normalized to obtain $\widetilde C$ as described in Appendix~\ref{app:ot-details}. With uniform marginals $\alpha_i=1/n_s$ and $\beta_j=1/n_t$, the entropically regularized plan is
\begin{equation}
\pi_q^\star
=
\underset{\pi\in\Pi(\alpha,\beta)}{\arg\min}
\left\{
\langle\pi,\widetilde C\rangle
+\varepsilon_q\sum_{i,j}\pi_{ij}(\log\pi_{ij}-1)
\right\}.
\label{eq:tempobridge-sinkhorn-plan}
\end{equation}
Sinkhorn iterations approximate the plan~\citep{cuturi2013sinkhorn}
without gradient tracking.
Endpoint pairs are sampled with replacement from its full joint distribution.

For a sampled pair $(z_0,z_1)$ and $\tau\sim\mathcal U(0,1)$, the training path and target velocity are
\begin{equation}
z_\tau=(1-\tau)z_0+\tau z_1,\qquad
u_\tau=z_1-z_0.
\label{eq:tempobridge-cfm-path}
\end{equation}
The flow-matching objective is
\begin{equation}
\mathcal L_{\mathrm{FM}}^{(q)}
=
\mathbb E_{(z_0,z_1)\sim\pi_q^\star,\tau}
\left[
\frac{1}{d_q}
\left\lVert
v_{\theta_q}(z_\tau,\tau,g_s,\eta_q)
-(z_1-z_0)
\right\rVert_2^2
\right].
\label{eq:tempobridge-flow-matching-loss}
\end{equation}
The expectation is estimated from sampled pairs across the training minibatch. OT pairs define training paths; they are not observed cell correspondences.

The objective for the transport-training stage combines flow matching, reconstruction of the two observed endpoints, and task-specific auxiliary terms:
\begin{equation}
\mathcal L_{\mathrm{total}}^{(q)}
=\lambda_{\mathrm{FM}}^{(q)}\mathcal L_{\mathrm{FM}}^{(q)}
+\lambda_{\mathrm{src}}^{(q)}\mathcal L_{\mathrm{rec,src}}^{(q)}
+\lambda_{\mathrm{tgt}}^{(q)}\mathcal L_{\mathrm{rec,tgt}}^{(q)}
+\sum_{k\in\mathcal A_q}\lambda_k^{(q)}\mathcal L_k^{(q)}.
\label{eq:tempobridge-total-loss}
\end{equation}
The reconstruction terms compare each observed endpoint with its own decoded latent representation, using a Huber loss~\citep{huber1964robust} that balances zero and nonzero expression entries. The set $\mathcal A_q$ contains the auxiliary terms enabled for task $q$.
Any representation or condition pretraining has a separate objective.

For differentiation, the auxiliary term orders predicted population means
along the direction between the observed endpoints. It uses full-interval
rollouts under virtual time descriptors; measured Day~1 and Day~2 expression
does not enter this loss. Intermediate prediction instead retains the
Day~0--Day~3 descriptor and changes the integration horizon.
For genetic and drug prediction, auxiliary terms constrain population
mean responses and latent or expression-space moments.
In the genetic objective, the population-mean and mean-shift correlation losses supervise quantities also assessed by Mean L2 and $\Delta r$.
Appendix~\ref{app:loss-details} specifies these terms and their coefficients.

\subsection{Population Generation}
\label{sec:tempobridge-population-generation}

Given source cells and a transition descriptor, the model computes
$Z_s$, $g_s$, and $\eta_q$, then integrates the neural ordinary differential equation~\citep{chen2018neural}:
\begin{equation}
\frac{\mathrm d z_i(\tau)}{\mathrm d\tau}
=
v_{\theta_q}(z_i(\tau),\tau,g_s,\eta_q),
\qquad
z_i(0)=\widetilde z_i^s,
\quad
0\leq\tau\leq r_q(c_q).
\label{eq:tempobridge-inference-ode}
\end{equation}
Both $g_s$ and $\eta_q$ remain fixed during integration.
The horizon is one for genetic and drug perturbations and follows the
fixed mapping above for temporal requests.
For temporal and genetic prediction, $\widetilde z_i^s=z_i^s$.
For drug prediction, $\widetilde z_i^s=z_i^s+\epsilon_i-\bar\epsilon$,
where the $\epsilon_i$ are independently sampled from $\mathcal N(0,I)$
and $\bar\epsilon$ is their mean within the generated population.
The number of output cells follows the task's inference protocol.
For each output position $i$, $(x_i^s,z_i^s)$ denotes the selected source
cell and its encoded latent state used for initialization.
Target expression and OT are not used at inference.

Let $\widehat z_i=z_i(r_q(c_q))$ denote the transported latent state at
the requested horizon.
Temporal prediction decodes the transported latent states as absolute
expression.
Genetic and drug prediction use source-anchored residual decoding:
\begin{equation}
\widehat x_i
=
\max\!\left\{
0,\,
x_i^s+D_q(\widehat z_i)-D_q(z_i^s)
\right\}.
\label{eq:tempobridge-source-anchored-decoding}
\end{equation}
The evaluation protocols specify source-cell selection, numerical
solvers, and checkpoint selection for each task.

\section{Experiments}
\label{sec:experiments}
\subsection{Experimental Protocol}
\label{sec:experimental-setup}

We evaluate population prediction for held-out donors, unseen perturbation components or combinations, and unseen compounds.
Controlled and matched comparisons assess source context and transport components. Within each task, methods share the evaluation split, gene space, sampling protocol, and evaluator.
The genetic feature panel was selected before condition splitting (Section~\ref{sec:gene-experiment}).
Complete comparisons appear in Appendix~\ref{app:complete-main-tables}.
Main comparisons use one validation-selected checkpoint per learned method and training setting, evaluated without retraining over three randomized test runs (seeds 42, 123, and 2026). Mean $\pm$ sample standard deviation (SD) summarizes run-level aggregates; genetic tables report means only. Retraining SD and donor- or compound-bootstrap intervals are reported separately. Drug main-table means weight conditions equally, whereas paired intervals weight compounds equally. Appendix~\ref{app:uncertainty-reporting} defines aggregation, statistical units, and uncertainty estimators.
Mean L2 and effect correlation $\Delta r$ measure population mean-response agreement; Energy distance~\citep{szekely2013energy} and maximum mean discrepancy (MMD)~\citep{gretton2012kernel} compare distributions. Differential-expression (DE) recovery uses top-response-gene overlap (DE@100) or fold-change rank correlation on reference DE genes (DE $\rho$). Drug pEnergy and pMMD use a common projected expression space. Task-specific estimators are defined in Appendix~\ref{app:metrics}.
\subsection{Differentiation in Held-Out Donors}
\label{sec:temporal-experiment}
The Cuomo iPSC time course~\citep{cuomo2020singlecell} tests whether the learned transport predicts intermediate populations for unseen donors. All cells from a donor remain in one partition. Representation pretraining and transition training use only training-donor Day~0 and Day~3 populations; measured Day~1/2 expression is excluded from both stages. Validation Day~0$\rightarrow$Day~3 predictions select checkpoints. Requiring at least 64 cells at Days~0, 1, and 2 yields 15 evaluation donors and 30 day~0$\rightarrow$Day~1/2 requests. For metric comparison, the evaluator sets the output count to the observed target count; target expression is not provided to the model.
TempoBridge achieved the lowest mean MMD and Energy and the highest mean $\Delta r$ and DE@100 among the primary evaluated methods (Table~\ref{tab:temporal-main-2009}). Energy was 0.1291 versus 0.1442 for scGen, and DE@100 was 0.5624 versus 0.5220 for PRESCIENT. Covariance recovery followed a different ranking: gene-covariance correlation was 0.4176 versus PRESCIENT's 0.4874. The auxiliary time-adapted scVIDR~\citep{kana2023scvidr} reference also had higher DE@100 ($0.694\pm0.003$; Appendix~\ref{app:scvidr-reference}). These comparisons assess full-model prediction of intermediate populations for held-out donors; Sections~\ref{sec:source-information} and~\ref{sec:ablation-analysis} examine the role of source context
separately.
\begin{table}[!htbp]
    \centering
    \caption{
    Temporal differentiation on 15 held-out donors, covering 30
    prediction requests in the 2,009-gene evaluation space.
    Scores are aggregated across donors and requested times within
    each run. Entries report mean $\pm$ standard deviation over three randomized test runs with fixed trained models.
    }
    \label{tab:temporal-main-2009}
    \small
    \setlength{\tabcolsep}{3pt}
    \renewcommand{\arraystretch}{1.1}
    \begin{tabular}{@{}lcccc@{}}
        \toprule
        Method
        & MMD $\downarrow$
        & Energy $\downarrow$
        & $\Delta r \uparrow$
        & DE@100 $\uparrow$ \\
        \midrule
        scGen
        & $0.4300 \pm 0.0072$
        & $0.1442 \pm 0.0019$
        & $0.6168 \pm 0.0049$
        & $0.5111 \pm 0.0043$ \\
        Conditional VAE~\citeyearpar{sohn2015learning}
        & $0.4243 \pm 0.0023$
        & $0.1462 \pm 0.0006$
        & $0.5826 \pm 0.0011$
        & $0.4532 \pm 0.0053$ \\
        Conditional MLP
        & $0.4302 \pm 0.0032$
        & $0.1507 \pm 0.0021$
        & $0.5742 \pm 0.0040$
        & $0.4448 \pm 0.0070$ \\
        CellOT
        & $0.4459 \pm 0.0105$
        & $0.1521 \pm 0.0022$
        & $0.5849 \pm 0.0053$
        & $0.4579 \pm 0.0082$ \\
        PRESCIENT
        & $0.4967 \pm 0.0052$
        & $0.1635 \pm 0.0035$
        & $0.6210 \pm 0.0064$
        & $0.5220 \pm 0.0035$ \\
        scNODE
        & $0.5628 \pm 0.0018$
        & $0.2294 \pm 0.0059$
        & $0.3543 \pm 0.0449$
        & $0.3316 \pm 0.0327$ \\
        TrajectoryNet
        & $0.5453 \pm 0.0023$
        & $0.2785 \pm 0.0042$
        & $-0.2845 \pm 0.0189$
        & $0.3484 \pm 0.0059$ \\
        \midrule
        \textbf{TempoBridge}
        & $\mathbf{0.3626 \pm 0.0143}$
        & $\mathbf{0.1291 \pm 0.0047}$
        & $\mathbf{0.6284 \pm 0.0115}$
        & $\mathbf{0.5624 \pm 0.0110}$ \\
        \bottomrule
    \end{tabular}
\end{table}

\subsection{Held-Out Genetic Perturbations}
\label{sec:gene-experiment}
The Norman K562 Perturb-seq benchmark~\citep{norman2019exploring} evaluates held-out intervention conditions using a fixed source panel of 128 control cells. It therefore tests perturbation prediction rather than the effect of varying source context. Seen~2/2 holds out combinations whose two components appear in training interventions; Seen~1/2 and Seen~0/2 expose one or neither component. Here ``unseen'' concerns intervention labels, not absence of the gene from expression inputs. Additive combines observed training single-gene responses and applies only to Seen~2/2. Seen~0/1 tests single genes held out in the Seen~0/2 split. The four test sets contain 20, 20, 16, and 11 conditions. Models are trained separately for the combination splits, with validation Mean L2 selecting checkpoints; Seen~0/1 reuses Seen~0/2 checkpoints without retraining or further selection. The gene panel includes highly variable genes selected from the full quality-controlled dataset before condition splitting (Appendix~\ref{app:task-inference-details}).
TempoBridge achieved Mean L2 2.2610 versus 3.1564 for scGPT-scratch~\citep{cui2024scgpt}, the lowest-L2 baseline; Additive obtained 3.6163 and $\Delta r=0.8719$, versus 0.9161 for TempoBridge (Table~\ref{tab:gene-by-setting-main}). Energy was 0.4793 versus 0.7335 for scDFM, the lowest-Energy baseline in this setting (Table~\ref{tab:gene-by-setting-full}). Across all three combination holdouts, TempoBridge had the lowest Mean L2 and Energy and highest $\Delta r$ among the evaluated methods. In Seen~0/1 it retained the lowest Mean L2 and highest $\Delta r$, while GEARS had lower Energy. Table~\ref{tab:gene-main} provides an auxiliary, equally weighted average of the four settings.
\begin{table}[!htbp]
\centering
\caption{Genetic prediction by holdout setting. Means use the test-repeat protocol in Section~\ref{sec:experimental-setup}; each setting weights its conditions equally. Additive applies only to Seen~2/2; NA denotes an inapplicable comparison. Bold marks the best mean within each setting and metric. Complete metrics are in Table~\ref{tab:gene-by-setting-full}.}
\label{tab:gene-by-setting-main}
\small
\setlength{\tabcolsep}{2.8pt}
\renewcommand{\arraystretch}{1.0}
\begin{tabular}{@{}l*{8}{c}@{}}
\toprule
& \multicolumn{2}{c}{Seen 2/2} & \multicolumn{2}{c}{Seen 1/2} & \multicolumn{2}{c}{Seen 0/2} & \multicolumn{2}{c}{Seen 0/1} \\
& \multicolumn{2}{c}{20 conditions} & \multicolumn{2}{c}{20 conditions} & \multicolumn{2}{c}{16 conditions} & \multicolumn{2}{c}{11 conditions} \\
\cmidrule(lr){2-3}\cmidrule(lr){4-5}\cmidrule(lr){6-7}\cmidrule(lr){8-9}
Method & L2$\downarrow$ & $\Delta r\uparrow$ & L2$\downarrow$ & $\Delta r\uparrow$ & L2$\downarrow$ & $\Delta r\uparrow$ & L2$\downarrow$ & $\Delta r\uparrow$ \\
\midrule
Control & 5.6384 & 0.0011 & 5.5351 & 0.0114 & 4.7860 & 0.0110 & 3.5682 & 0.0113 \\
Additive & 3.6163 & 0.8719 & \multicolumn{2}{c}{NA} & \multicolumn{2}{c}{NA} & \multicolumn{2}{c}{NA} \\
scGPT-scratch & 3.1564 & 0.8635 & 4.1885 & 0.6614 & 4.1235 & 0.5237 & 3.4632 & 0.4203 \\
GEARS & 3.8881 & 0.7330 & 4.7681 & 0.4596 & 4.2786 & 0.5081 & 3.4356 & 0.3338 \\
CPA & 5.0754 & 0.4594 & 5.3329 & 0.3943 & 6.1414 & 0.2933 & 4.1867 & 0.3266 \\
CellFlow & 3.6758 & 0.8263 & 4.4758 & 0.6697 & 4.1562 & 0.5436 & 3.2767 & 0.4236 \\
scDFM & 3.7269 & 0.8964 & 4.7935 & 0.6095 & 5.5934 & 0.5014 & 3.9321 & 0.2359 \\
TempoBridge & \textbf{2.2610} & \textbf{0.9161} & \textbf{3.9962} & \textbf{0.6741} & \textbf{3.9880} & \textbf{0.5589} & \textbf{3.2307} & \textbf{0.4330} \\
\bottomrule
\end{tabular}
\end{table}
Mean-response and Energy improvements were more consistent across combination holdouts than DE ranking, whose strongest method varied by setting (Table~\ref{tab:gene-by-setting-full}). In the equally weighted four-setting summary, DE $\rho$ was 0.6628 for TempoBridge versus 0.7474 for scGPT-scratch; TempoBridge led Seen~2/2 but ranked below the other trained methods in Seen~0/1. These results evaluate the complete training recipe, including auxiliary supervision, without isolating OT or flow contributions. Prediction with the ID-based descriptor also does not establish functional identification of unseen genes (Appendices~\ref{app:task-parameterization} and~\ref{app:loss-details}).
\subsection{Response Prediction for Held-Out Compounds}
\label{sec:drug-experiment}
The compound-disjoint sci-Plex3 split~\citep{srivatsan2020massively} contains 150 training, 19 validation, and 19 test compounds across A549, K562, and MCF7 cells. Predictions start from vehicle controls matched by cell line and replicate. Replicates are pooled within cell-line--compound--dose--time conditions, and metrics weight the 248 test conditions equally. This task tests molecular descriptor conditioning for compounds absent from training and validation, without a learned training-compound identity.
chemCPA, scDFM, and CellFlow~\citep{hetzel2022chemcpa,yu2026scdfm,klein2025cellflow} are evaluated through project-specific adapters with different conditioning and output interfaces (Appendix~\ref{app:drug-adapter-interface}). Their expression predictions enter the same evaluator; treatment duration is not an explicit input to the chemCPA adapter. CellFlow's internal, train-fitted PCA is distinct from the common 50-dimensional Gaussian projection used for pEnergy and pMMD (seed 42); MMD bandwidth is calibrated on training populations. These comparisons concern the implemented adapters under this protocol.
With conditions weighted equally, TempoBridge had the lowest Mean L2 and pEnergy and the highest $\Delta r$ among the evaluated methods (Table~\ref{tab:drug-main}).
Mean L2 was 4.2478 versus 4.2875 for the CellFlow adapter, and $\Delta r$ was 0.5977 versus 0.5608 for the same comparator.
pEnergy was 0.2123 versus 0.2219 for CFM without OT and 0.2260 for Conditional MLP.
In a separate compound-weighted paired analysis, the 95\% confidence intervals for the pEnergy differences against CFM without OT and Conditional MLP both included zero (Appendix~\ref{app:drug-evidence}).
TempoBridge ranked second on pMMD (0.0327 versus 0.0317 for Conditional MLP) and DE@100 (0.5471 versus 0.5579 for the scDFM adapter).
\begin{table}[!htbp]
\centering
\caption{Unseen-compound response prediction on sci-Plex3 (248 equally weighted test conditions). Values are mean $\pm$ sample SD over fixed-checkpoint test repeats (Section~\ref{sec:experimental-setup}). pEnergy and pMMD use a common 50-dimensional Gaussian random projection (seed 42), with MMD bandwidth calibrated on training data. Baseline adapters are project implementations; bold marks the best mean in each column.}
\label{tab:drug-main}
\footnotesize
\setlength{\tabcolsep}{2.5pt}
\begin{tabular}{@{}lccccc@{}}
\toprule
Method & Mean L2$\downarrow$ & $\Delta r\uparrow$ & pEnergy$\downarrow$ & pMMD$\downarrow$ & DE@100$\uparrow$ \\
\midrule
Conditional MLP & 4.8777$\pm$0.0455 & 0.2254$\pm$0.0190 & 0.2260$\pm$0.0095 & \textbf{0.0317$\pm$0.0005} & 0.3215$\pm$0.0228 \\
CFM without OT & 4.6357$\pm$0.0285 & 0.3848$\pm$0.0056 & 0.2219$\pm$0.0094 & 0.0345$\pm$0.0027 & 0.2626$\pm$0.0054 \\
chemCPA adapter & 5.5117$\pm$0.1670 & 0.3827$\pm$0.0122 & 0.3484$\pm$0.0358 & 0.0437$\pm$0.0029 & 0.5139$\pm$0.0098 \\
scDFM adapter & 7.7612$\pm$0.5215 & 0.2293$\pm$0.0220 & 0.8281$\pm$0.1396 & 0.0747$\pm$0.0079 & \textbf{0.5579$\pm$0.0034} \\
CellFlow adapter & 4.2875$\pm$0.0125 & 0.5608$\pm$0.0002 & 9.2228$\pm$0.0121 & 0.2714$\pm$0.0001 & 0.3465$\pm$0.0002 \\
\midrule
\tb & \textbf{4.2478$\pm$0.0470} & \textbf{0.5977$\pm$0.0047} & \textbf{0.2123$\pm$0.0193} & 0.0327$\pm$0.0036 & 0.5471$\pm$0.0129 \\
\bottomrule
\end{tabular}
\end{table}

\subsection{Source-Context Information and Model Use}
\label{sec:source-information}
The three task comparisons do not isolate source context. We next test its information content, its use by a fitted model, and its predictive benefit under matched retraining.

\subsubsection{Controlled Source Ambiguity}
\label{sec:controlled-source-context}

We tested the information restriction in Section~\ref{sec:source-ambiguity} using two synthetic source populations with exactly the same central cells and transition descriptor. Additional cells on the left or right distinguished the populations; their unpaired target populations required the central cells to reach opposite vertical branches. A parameter-matched no-context model replaced the pooled context with zeros. With identical shared initial states and descriptors, its deterministic field and solver produce identical shared-cell outputs for both populations. The additional source cells therefore supply the information needed to distinguish the two requests.
Across three test runs of one fixed checkpoint per trained arm, branch accuracy was 50.00\% without context, 98.04\% with correct context, and 1.96\% with swapped context (Table~\ref{tab:p0-1-main}). Correct context also gave the lowest full-population Energy in this comparison. Appendix~\ref{app:controlled-source-context} gives the construction, evaluation protocol, and controls.
\begin{table}[!htbp]
    \centering
    \caption{
Controlled source-dependent transition ambiguity.
Results are mean $\pm$ sample SD across three test runs of one fixed checkpoint per trained arm.
Branch accuracy evaluates shared synthetic cells; Energy evaluates the full population.
Correct and swapped contexts use the same TempoBridge checkpoint.
}
    \label{tab:p0-1-main}
    \small
    \setlength{\tabcolsep}{8pt}
    \renewcommand{\arraystretch}{1.0}
    \begin{tabular}{@{}lcc@{}}
        \toprule
        Model or setting
        & Branch acc.\ (\%) $\uparrow$
        & Energy $\downarrow$ \\
        \midrule
        No context
        & $50.00 \pm 0.00$
        & $0.4575 \pm 0.0012$ \\
        Swapped context
        & $1.96 \pm 0.40$
        & $3.3299 \pm 0.1760$ \\
        TempoBridge
        & $98.04 \pm 0.40$
        & $0.0054 \pm 0.0004$ \\
        \bottomrule
    \end{tabular}
\end{table}

\subsubsection{Donor-Context Swaps}
\label{sec:biological-source-context}
Using the held-out donors from Section~\ref{sec:temporal-experiment}, we split each donor's Day~0 cells into disjoint query and context subsets to test source-context use. With query cells, requested day, model parameters, and solver fixed, we replaced only the pooled context with each of the other 14 donors' contexts. Averaging incorrect contexts increased Energy by 0.00424 (donor-bootstrap 95\% CI $[0.00180,0.00678]$), with 12 of 15 donors favoring the correct context (Table~\ref{tab:temporal-context-swap}). MMD increased by 0.00657 and $\Delta r$ decreased by 0.01124, with intervals also excluding zero. This intervention indicates use of empirical source information by the fitted model (Appendix~\ref{app:temporal-context-swap}); Section~\ref{sec:ablation-analysis} separately tests predictive benefit under matched retraining.
\subsection{Transport and Conditioning Ablations}
\label{sec:ablation-analysis}
\paragraph{Temporal transport variants.}
Temporal variants replace OT coupling with independent random pairing, or retain OT endpoint sampling while learning a single latent residual update. Random pairing increased mean Energy from 0.1291 to 0.1446; static OT residual increased it to 0.1350 (Table~\ref{tab:temporal-ablation-2009}). Both variants also reduced $\Delta r$ and DE@100 and increased MMD. The static formulation replaces pathwise velocity regression with residual regression and omits the flow-specific ordering term (Appendix~\ref{app:temporal-ordering}). These results favor the full temporal training formulation; the static comparison does not isolate numerical integration alone.
\paragraph{Source context with matched retraining.}
We retrained temporal models with and without source context to test predictive benefit separately from the fixed-model swaps in Section~\ref{sec:biological-source-context}. Across three training seeds, the context model had lower mean MMD (0.378 vs. 0.397) and Energy (0.135 vs. 0.139), while the no-context model had higher $\Delta r$ (0.639 vs. 0.625) and DE@100 (0.529 vs. 0.515; Table~\ref{tab:source-context-ablation-2009}). Context therefore favored distributional distances, whereas no context favored effect correlation and DE@100.
\paragraph{Drug-conditioning variants.}
A matched $2\times2$ Stage-B analysis varies source-context and cell-line conditioning while sharing a frozen expression backbone and condition warm-start (Table~\ref{tab:drug-context-ablation}). For Full minus No source context, the compound-weighted paired effects were 0.004 for $\Delta r$ (95\% CI $[0.001,0.006]$) and 0.019 for DE $\rho$ ($[0.008,0.033]$). Under condition weights, pEnergy slightly favored No source context and pMMD was identical at the reported precision; removing cell-line conditioning reduced $\Delta r$ from 0.425 to 0.414. These metric-specific findings concern Stage-B conditioning, separately from the main drug comparison and its test-repeat variability (Appendix~\ref{app:drug-evidence}).
\section{Discussion and Conclusion}
\label{sec:conclusion}

The controlled construction identifies an information restriction when
local state and the requested transition are identical. Frozen-model
donor swaps show that the fitted field uses source context on biological
data. Matched retraining separately tests whether supplying that
information improves prediction; full-model comparisons alone do not
isolate the contribution of source context.

Predictive benefit depends on the setting and metric. In matched temporal
retraining, context favored distributional distances, whereas the
no-context model favored effect correlation and DE@100; the drug
conditioning analysis also showed metric tradeoffs. The fixed-source
genetic benchmark evaluates conditional perturbation prediction rather
than source-context benefit. Across the biological tasks, mean-response
and population-distribution gains were more consistent than covariance
recovery and DE ranking.

TempoBridge addresses missing source-population information in unpaired
transition prediction by conditioning the transport field separately on
evolving cell states, a fixed initial empirical source-population summary,
and the requested transition. Temporal, genetic, and drug models are
trained independently. OT couplings construct the flow-matching training
paths, while inference requires only the source population and transition
descriptor.
\subsection*{AI use statement}

Generative AI tools assisted with manuscript organization, language editing,
and code development and debugging for method implementation, experiments,
and figure formatting.
Numerical figures were prepared from existing experimental records;
these tools were not used to generate experimental observations.
The authors take responsibility for the final text, claims, code, and other
artifacts, including all material produced with AI assistance.

\subsection*{Ethics statement}

This study retrospectively analyzes previously collected single-cell
population measurements.
We do not claim individual-level clinical prediction, biological lineage
identification, or causal effects of interventions.
Dataset access and reuse follow the conditions specified by the original
data providers.

\subsection*{Reproducibility statement}

The main text and appendix document task-specific splits, target-visibility
boundaries, model settings, model-selection rules, and evaluation estimators.
An anonymized code archive is provided as supplementary material.
The reported comparisons are retrospective evaluations, as described in
Section~\ref{sec:experimental-setup}.

Each principal learned comparison uses one checkpoint selected by validation
metrics per method and training setting.
The selected checkpoint is evaluated using test random seeds 42, 123, and
2026, with no retraining or changes to its weights between runs.
Reported means and, where applicable, sample standard deviations (SDs) are
calculated from the three run-level aggregate scores.
These SDs describe test-run variation conditional on the selected fitted
models.
Appendix~\ref{app:uncertainty-reporting} defines the aggregation procedures
and sample SD estimator.

The genetic tables report means only.
Table~\ref{tab:gene-by-setting-main} reports each setting separately, while
Table~\ref{tab:gene-main} assigns equal weight to the four setting-level means.
Genetic predictions use one fixed exponential moving average (EMA) parameter
set per trained model, without combining predictions across training seeds.

The drug main table evaluates trained two-stage pipelines under the test-run
protocol described above.
Drug pMMD and pEnergy use a fixed Gaussian random projection, with the MMD
bandwidth calibrated on training data.
The fixed-backbone Stage-B training-seed analysis is reported separately
in the appendix.
The Gaussian and PCA sensitivity analyses in
Appendix~\ref{app:drug-evidence} also concern the fixed-backbone setting.
Project-specific adapters are identified in the corresponding drug comparison.

The controlled source-context experiment uses one fixed checkpoint per
trained arm and three test runs. Supplementary analyses based on independently
trained models report training-seed variability separately from these
fixed-checkpoint test-run SDs.
Projection seeds define fixed transforms and are not statistical replicates.
Paired bootstrap intervals describe variation across biological units
conditional on the fitted models and are reported separately from both
test-run and training-seed SDs.
Drug main-table means weight conditions equally, whereas the paired
intervals in Appendix~\ref{app:drug-evidence} concern a comparison that
weights compounds equally.

\bibliography{references}
\bibliographystyle{plainnat}
\clearpage
\appendix
\setcounter{figure}{0}
\renewcommand{\thefigure}{S\arabic{figure}}
\setcounter{table}{0}
\renewcommand{\thetable}{S\arabic{table}}
\renewcommand{\theHtable}{supp.\arabic{table}}
\section{Complete Task and Prediction Results}
\label{app:complete-main-tables}
The tables below retain the complete comparisons underlying the main-text figures, including all reported metrics and all evaluated methods. Table~\ref{tab:task-interface} summarizes the task interfaces. Figure panels show selected metrics; claims about other metrics refer to the corresponding tables.
\begin{table}[!htbp]
\centering
\caption{
Task-specific instances of the conditional population prediction
problem. Each task uses an independently trained model and its
own gene space.
The tasks share a common formulation,
but not model parameters.
}
\label{tab:task-interface}

\small
\setlength{\tabcolsep}{5pt}
\renewcommand{\arraystretch}{1.0}

\begin{tabular}{
    @{}
    >{\raggedright\arraybackslash}p{0.17\linewidth}
    >{\raggedright\arraybackslash}p{0.20\linewidth}
    >{\raggedright\arraybackslash}p{0.27\linewidth}
    >{\raggedright\arraybackslash}
        p{\dimexpr0.36\linewidth-6\tabcolsep\relax}
    @{}
}
    \toprule
    \textbf{Task}
    & \textbf{Source population}
    & \textbf{Transition descriptor}
    & \textbf{Target population and evaluation setting} \\
    \midrule

    Temporal differentiation
    & Day~0 cells from a held-out donor
    & Requested target time
    & Day~1 or Day~2 population of the same donor \\

    \addlinespace[6pt]

    Genetic perturbation
    & Fixed K562 control population
    & Unordered set of one or two perturbed genes
    & Perturbed population under controlled component exposure \\

    \addlinespace[6pt]

    Drug response
    & Matched vehicle-control population
    & Molecular description, dose, treatment time, and cell line
    & Treated population for a compound excluded from
      training and validation \\

    \bottomrule
\end{tabular}
\end{table}

\begin{table}[!htbp]
\centering
\caption{Complete genetic metrics by holdout setting. For learned methods, each entry is the mean across test seeds 42, 123, and 2026 using the fixed checkpoint for that training setting, after condition-level averaging within each test run. Control and Additive are deterministic. Additive uses the observed single-gene responses and applies only to Seen~2/2. NA denotes an undefined metric for Control or an inapplicable comparison for Additive. Bold marks the best mean for each setting and metric. Seen~0/1 reuses the Seen~0/2 checkpoints.
Control copies a target-sized prefix of the fixed source panel; its effect
correlations exclude undefined values, as detailed in Appendix~\ref{app:metrics}.}
\label{tab:gene-by-setting-full}
\small
\setlength{\tabcolsep}{7pt}
\renewcommand{\arraystretch}{1.0}
\begin{tabular}{@{}lcccc@{}}
\toprule
Method & Mean L2$\downarrow$ & $\Delta r\uparrow$ & Energy$\downarrow$ & DE $\rho\uparrow$ \\
\midrule
\multicolumn{5}{l}{Seen~2/2: 20 test conditions} \\
\midrule
Control & 5.6384 & 0.0011 & 1.8371 & 0.0378 \\
Additive & 3.6163 & 0.8719 & 0.9871 & 0.8956 \\
scGPT-scratch & 3.1564 & 0.8635 & 7.8094 & 0.9034 \\
GEARS & 3.8881 & 0.7330 & 0.9807 & 0.7925 \\
CPA & 5.0754 & 0.4594 & 7.2051 & 0.6198 \\
CellFlow & 3.6758 & 0.8263 & 2.7345 & 0.8584 \\
scDFM & 3.7269 & 0.8964 & 0.7335 & 0.9160 \\
TempoBridge & \textbf{2.2610} & \textbf{0.9161} & \textbf{0.4793} & \textbf{0.9233} \\
\midrule
\multicolumn{5}{l}{Seen~1/2: 20 test conditions} \\
\midrule
Control & 5.5351 & 0.0114 & 1.7694 & 0.0900 \\
Additive & NA & NA & NA & NA \\
scGPT-scratch & 4.1885 & 0.6614 & 8.4925 & \textbf{0.7678} \\
GEARS & 4.7681 & 0.4596 & 1.3841 & 0.5816 \\
CPA & 5.3329 & 0.3943 & 7.7166 & 0.5833 \\
CellFlow & 4.4758 & 0.6697 & 3.2712 & 0.7460 \\
scDFM & 4.7935 & 0.6095 & 1.2584 & 0.7474 \\
TempoBridge & \textbf{3.9962} & \textbf{0.6741} & \textbf{1.0692} & 0.6978 \\
\midrule
\multicolumn{5}{l}{Seen~0/2: 16 test conditions} \\
\midrule
Control & 4.7860 & 0.0110 & 1.3573 & 0.0196 \\
Additive & NA & NA & NA & NA \\
scGPT-scratch & 4.1235 & 0.5237 & 8.0856 & 0.6873 \\
GEARS & 4.2786 & 0.5081 & 1.0833 & \textbf{0.6927} \\
CPA & 6.1414 & 0.2933 & 8.5650 & 0.5235 \\
CellFlow & 4.1562 & 0.5436 & 2.9558 & 0.6456 \\
scDFM & 5.5934 & 0.5014 & 1.8711 & 0.6637 \\
TempoBridge & \textbf{3.9880} & \textbf{0.5589} & \textbf{0.9738} & 0.6102 \\
\midrule
\multicolumn{5}{l}{Seen~0/1: 11 test conditions} \\
\midrule
Control & 3.5682 & 0.0113 & 0.7880 & NA \\
Additive & NA & NA & NA & NA \\
scGPT-scratch & 3.4632 & 0.4203 & 7.3042 & 0.6311 \\
GEARS & 3.4356 & 0.3338 & \textbf{0.7118} & \textbf{0.6725} \\
CPA & 4.1867 & 0.3266 & 6.5700 & 0.5276 \\
CellFlow & 3.2767 & 0.4236 & 2.2575 & 0.5856 \\
scDFM & 3.9321 & 0.2359 & 0.9585 & 0.4509 \\
TempoBridge & \textbf{3.2307} & \textbf{0.4330} & 0.7133 & 0.4198 \\
\bottomrule
\end{tabular}
\end{table}

\begin{table}[!htbp]
    \centering
    \caption{
    Genetic perturbation prediction averaged over four settings:
    Seen~2/2 (two seen genes; 20 conditions),
    Seen~1/2 (one seen and one unseen gene; 20 conditions),
    Seen~0/2 (two unseen genes; 16 conditions), and
    Seen~0/1 (one unseen gene; 11 conditions).
    For each method and metric, the reported value is the arithmetic mean
    of the four setting-level scores, giving each setting equal weight.
    Each setting-level score first averages test conditions within each run,
    then averages test seeds 42, 123, and 2026 using one fixed checkpoint per
    learned method and training setting. Only means are reported; no
    cross-setting SD is computed.
    DE $\rho$ denotes Spearman correlation between predicted and observed
    log-fold changes on reference DE genes.
    Control is deterministic; NA indicates an undefined four-setting average.
    Additive is reported in Table~\ref{tab:gene-by-setting-full} for Seen~2/2;
    its four-setting average is not defined.
    Seen~0/1 reuses the Seen~0/2 checkpoints without retraining.
    }
    \label{tab:gene-main}
    \small
    \setlength{\tabcolsep}{5pt}
    \renewcommand{\arraystretch}{1.04}
    \begin{tabular}{@{}lcccc@{}}
        \toprule
        Method
        & Mean L2 $\downarrow$
        & $\Delta r \uparrow$
        & Energy $\downarrow$
        & DE $\rho \uparrow$ \\
        \midrule
        Control
        & 4.8819 & 0.0087 & 1.4380 & NA \\
        scGPT-scratch
        & 3.7329 & 0.6172 & 7.9229 & \textbf{0.7474} \\
        GEARS
        & 4.0926 & 0.5086 & 1.0400 & 0.6848 \\
        CPA
        & 5.1841 & 0.3684 & 7.5142 & 0.5636 \\
        CellFlow
        & 3.8961 & 0.6158 & 2.8048 & 0.7089 \\
        scDFM
        & 4.5115 & 0.5608 & 1.2054 & 0.6945 \\
        \tb
        & \textbf{3.3690}
        & \textbf{0.6455}
        & \textbf{0.8089}
        & 0.6628 \\
        \bottomrule
    \end{tabular}
\end{table}

\begin{table}[!htbp]
    \centering
    \caption{
    Temporal transport-formulation comparisons under the same donor,
    gene-space, and evaluation protocol as
    Table~\ref{tab:temporal-main-2009}.
    Entries report mean $\pm$ sample SD over three randomized test runs of
    one fixed checkpoint per variant (test seeds 42, 123, and 2026),
    using the same run-level aggregation
    as the temporal main comparison.
    }
    \label{tab:temporal-ablation-2009}
    \small
    \setlength{\tabcolsep}{3pt}
    \renewcommand{\arraystretch}{1.1}
    \begin{tabular}{@{}lcccc@{}}
        \toprule
        Variant
        & MMD $\downarrow$
        & Energy $\downarrow$
        & $\Delta r \uparrow$
        & DE@100 $\uparrow$ \\
        \midrule
        Random-pairing flow
        & $0.4145 \pm 0.0340$
        & $0.1446 \pm 0.0063$
        & $0.6080 \pm 0.0102$
        & $0.5316 \pm 0.0116$ \\
        Static OT residual
        & $0.3848 \pm 0.0207$
        & $0.1350 \pm 0.0049$
        & $0.6159 \pm 0.0105$
        & $0.5071 \pm 0.0105$ \\
        \textbf{Full flow + OT}
        & $\mathbf{0.3626 \pm 0.0143}$
        & $\mathbf{0.1291 \pm 0.0047}$
        & $\mathbf{0.6284 \pm 0.0115}$
        & $\mathbf{0.5624 \pm 0.0110}$ \\
        \bottomrule
    \end{tabular}
\end{table}

\setcounter{figure}{0}
\renewcommand{\thefigure}{S\arabic{figure}}
\renewcommand{\theHfigure}{supp.\arabic{figure}}
\section{Detailed Task Parameterization}
\label{sec:appendix}
\label{app:task-parameterization}

The main comparisons follow the protocol in
Section~\ref{sec:experimental-setup}: each learned method and training setting
contributes one selected checkpoint, evaluated with test random seeds 42,
123, and 2026 while its parameters remain fixed.
Training-seed identifiers describe how checkpoints were obtained; test-repeat
identifiers index repeated evaluation of a selected checkpoint. Supplementary
analyses that aggregate independently trained checkpoints state this separately.
The training-seed records below therefore do not define the three main-table
test repeats. Projection seeds specify fixed evaluation or loss transforms and
are not statistical replicates. Appendix~\ref{app:uncertainty-reporting} gives
the aggregation formulas.

\subsection{Factorized conditioning and source context}
\label{app:context-parameterization}

The descriptor $c_q$ contains the structured information needed to identify a
requested target condition and is implemented separately for each task. In
every TempoBridge instance, the
descriptor encoding remains separate from the context computed from the
observed source cells. Let $E_{\phi_q}$ be the task-specific expression encoder
and let $m_i$ mark valid cells in a padded population batch. The source context
used by the velocity field is
\begin{equation}
    h(\widehat P_s)=\bar z_s
    =\frac{1}{\sum_i m_i}\sum_i m_iE_{\phi_q}(x_i^s).
    \label{app:eq:source-context}
\end{equation}
Pooling uses the valid source cells in the input population record, before
OT-specific endpoint sampling; the record can itself be sampled by the data
loader. OT subsampling retains this summary rather than recomputing it from
the selected endpoints. For donor-context swaps, a disjoint context subset
supplies the summary, as specified in Appendix~\ref{app:temporal-context-swap}.
This mean-pooled latent vector is a global context, not a complete encoding of
the source distribution; cell-level variation remains in the individual
transported latent states. The pooling operation is permutation invariant. For
any permutation $\pi$ of the valid source cells,
\begin{equation}
    h\!\left(\{x_{\pi(i)}^s\}_{i=1}^{n_s}\right)
    =h\!\left(\{x_i^s\}_{i=1}^{n_s}\right).
    \label{app:eq:permutation-invariance}
\end{equation}

Table~\ref{app:tab:conditioning-audit} distinguishes empirical population
context, explicit descriptor fields, and metadata that is not supplied to the
predictor. Dataset and task identifiers may appear as implementation
tokens, but they are constant within the separately trained task instances and
do not create a shared-parameter multitask model.

\begin{table}[!htbp]
    \centering
    \caption{Conditioning interface used by the three task instances.}
    \label{app:tab:conditioning-audit}
    \footnotesize
    \setlength{\tabcolsep}{4pt}
    \renewcommand{\arraystretch}{1.12}
    \begin{tabular}{@{}>{\raggedright\arraybackslash}p{0.15\linewidth}
                       >{\raggedright\arraybackslash}p{0.24\linewidth}
                       >{\raggedright\arraybackslash}p{0.34\linewidth}
                       >{\raggedright\arraybackslash}p{0.19\linewidth}@{}}
        \toprule
        Task & Context from source cells & Structured descriptor fields
        & Not supplied as predictor input \\
        \midrule
        Temporal
        & Day~0 population from the requested donor
        & Endpoint times; requested time through the integration horizon
        & Donor identity; replicate label \\
        \addlinespace[3pt]
        Genetic
        & Matched K562 control population
        & Perturbed genes; control/single/combination type
        & Combination identity hash; replicate label \\
        \addlinespace[3pt]
        Drug
        & Matched vehicle-control population
        & Molecular feature, dose, treatment time, cell line
        & Drug identity in the unseen-compound setting; replicate label \\
        \bottomrule
    \end{tabular}
\end{table}

Replicate labels are used to construct populations and check for leakage, not to
specify a requested biological transition.

\subsection{Temporal descriptor and integration request}
\label{app:temporal-parameterization}

At the formulation level, the temporal descriptor contains an endpoint
transition and a requested time,
\begin{equation}
    c_{\mathrm{time}}=(c_{\mathrm{end}},t_r),
    \qquad
    c_{\mathrm{end}}=(t_s,t_e,\Delta t),
    \qquad
    \Delta t=t_e-t_s.
    \label{app:eq:temporal-descriptor}
\end{equation}
The temporal instance implements these two parts through different
interfaces. For endpoint flow matching and evaluation, the condition encoder receives the
fixed descriptor $c_{\mathrm{end}}=(0,3,3)$, whereas $t_r$ determines how far
the endpoint-conditioned field is integrated. The auxiliary ordering branch
uses virtual time descriptors. Appendix~\ref{app:temporal-ordering}
specifies the request-to-horizon mapping, ordering regularizer,
target-visibility boundary, and numerical solver.

\subsection{Genetic-perturbation descriptor}
\label{app:gene-parameterization}

For the Norman instance, the descriptor is
\begin{equation}
    c_{\mathrm{gene}}=(\mathcal G,\rho),
    \qquad
    \mathcal G=\{g_k\}_{k=1}^{K},
    \qquad
    \rho\in\{\mathrm{control},\mathrm{single},\mathrm{combination}\}.
    \label{app:eq:gene-descriptor}
\end{equation}
Here $K=1$ for a single-gene intervention and $K=2$ for the
combinatorial conditions. Gene identifiers are mapped to the expression gene
space, padded positions contribute no mass, and the component embeddings are
mean-pooled,
\begin{equation}
    e_{\mathcal G}=\frac{1}{K}\sum_{g\in\mathcal G}e_g.
    \label{app:eq:gene-composition}
\end{equation}
The pooled representation is combined with a learned intervention-type token.
The condition encoder uses a separate 512-dimensional gene-embedding table,
initialized from a zero-mean normal distribution with standard deviation 0.02;
the ObjectiveEMA configuration does not load pretrained gene embeddings.
An unseen gene here is absent from training intervention labels, not from
the expression input space. Its embedding row receives no intervention-specific
supervision. Shared downstream parameters are learned
from training conditions, and optimizer weight decay can still change that row.
Permutation invariance therefore does not imply an external functional prior
for an unseen gene.
This construction is invariant to the order of genes in a combination and does
not use a combination-identity hash. A dedicated control token represents the
unperturbed condition. Target expression for held-out combinations is not an
inference input.

\subsection{Drug-response descriptor}
\label{app:drug-parameterization}

For the compound-disjoint sci-Plex3 instance, the task descriptor is
\begin{equation}
    c_{\mathrm{drug}}=(f_m,a,t_d,\ell),
    \qquad f_m\in\mathbb{R}^{256},
    \label{app:eq:drug-descriptor}
\end{equation}
where $f_m$ is a SMILES-derived character $n$-gram representation, $a$ is the
dose, $t_d$ is treatment time, and $\ell$ is the cell-line category encoded
from the training vocabulary. The molecular representation is not a
Morgan/ECFP fingerprint. Because validation and evaluation compounds are
disjoint from training compounds, the discrete drug-identity embedding is
disabled. Compounds absent from training are represented through $f_m$ and the
remaining structured fields. Cell-line metadata in $c_{\mathrm{drug}}$ is distinct from
$h(\widehat P_s)$, which is computed from the actual matched vehicle-control
cells. Replicate labels are used for grouping rather than conditioning. The
two-stage training and source-anchored decoding rules are given in
Appendix~\ref{app:task-inference-details}.

\subsection{Expected risk under source-dependent ambiguity}
\label{app:source-risk}

Fix the distribution of training records, couplings, paths, and flow time. Let $S$ denote the source population, $X=(Z_\tau,\tau,C)$, and $U=U_\tau$, with $\mathbb E\lVert U\rVert_2^2<\infty$. Define $R^*(V)=\inf_f\mathbb E\lVert U-f(V)\rVert_2^2$, where the infimum is over all measurable square-integrable predictors. Writing $m(X,S)=\mathbb E[U\mid X,S]$ and $\bar m(X)=\mathbb E[U\mid X]$ gives
\[
R^*(X)-R^*(X,S)=\mathbb E\lVert m(X,S)-\bar m(X)\rVert_2^2.
\]
To prove the identity, expand $U-\bar m=(U-m)+(m-\bar m)$. The cross term vanishes after conditioning on $(X,S)$, and conditional means minimize squared error. Thus the risk gap is positive exactly when $m(X,S)\ne\bar m(X)$ with positive probability.

For a quantitative bound, suppose $S\in\{0,1\}$, let $w(X)=\Pr(S=1\mid X)$, and write $m_j(X)=\mathbb E[U\mid X,S=j]$. Assume an $X$-measurable event $A$ has probability $p>0$ and, almost surely on $A$, $a\leq w(X)\leq1-a$ and $\lVert m_1(X)-m_0(X)\rVert_2\geq\delta$, where $0<a\leq1/2$ and $\delta>0$. Then
\[
\begin{aligned}
R^*(X)-R^*(X,S)
&=\mathbb E\!\left[w(X)(1-w(X))\lVert m_1(X)-m_0(X)\rVert_2^2\right]\\
&\geq p\,a(1-a)\delta^2.
\end{aligned}
\]
The equality is the conditional variance of a two-point vector; restricting its expectation to $A$ proves the bound. Divide by latent dimension for dimension-normalized FM loss.

For a deterministic summary $H=h(S)$, the same argument gives
\[
R^*(X,H)-R^*(X,S)
=\mathbb E\lVert m(X,S)-\mathbb E[U\mid X,H]\rVert_2^2.
\]
Consequently, a summary attains the full-source optimum exactly when these conditional means agree almost surely. These claims concern the fixed path law; shared initial states in the synthetic construction alone do not establish the required positive-probability path overlap.

\section{TempoBridge Implementation Details}
\label{app:implementation-details}

\subsection{Expression encoders and decoders}
\label{app:expression-models}

Each task uses its own expression encoder and decoder. The temporal and genetic
encoders remain trainable during transition learning. The drug instance first
trains an expression-reconstruction backbone on the Task~3 training split and
then freezes the source encoder and gene embedding during transition training;
its decoder remains trainable. The temporal decoder predicts an absolute
expression state, whereas the genetic and drug decoders predict
source-anchored expression changes. Final expression predictions are constrained to be nonnegative. Gene-space and latent dimensions and the
corresponding inference settings are listed in
Table~\ref{app:tab:task-inference}. Appendix~\ref{app:model-details} gives the
full temporal architecture and optimization record.

\subsection{Descriptor encoders}
\label{app:descriptor-details}

All task instances preserve the factorization introduced in
Section~\ref{sec:tempobridge}: source context is computed from observed
source cells, whereas the descriptor encoder receives structured information
about the target condition. The temporal instance encodes the Day~0--Day~3
endpoint and supplies the requested day through the integration horizon. The
genetic instance mean-pools the embeddings of the requested genes and combines
the result with an intervention-type token. The drug instance encodes a
256-dimensional SMILES character $n$-gram vector, dose, treatment
time, and cell line. Its drug-identity embedding is disabled for
compound-disjoint evaluation. Appendix~\ref{app:task-parameterization} gives the
exact descriptor tuples and the metadata excluded from prediction.
For batched computation, populations are padded and a validity mask excludes
padded positions from the latent mean used as source context.

\subsection{OT coupling}
\label{app:ot-details}

For numerical stability, squared latent distances are normalized within each
minibatch:
\begin{equation}
    M_{ij}=
    \frac{\lVert z_i^s-z_j^t\rVert_2^2}
    {\operatorname{median}_{i,j}\lVert z_i^s-z_j^t\rVert_2^2}.
    \label{app:eq:ot-cost}
\end{equation}
The entropic OT problem used to construct the regression paths uses uniform
marginals, after which endpoint pairs are sampled from the joint plan. The
temporal protocol uses $\varepsilon=0.03$ and 200 Sinkhorn iterations;
the genetic protocol uses $\varepsilon=0.05$ and 30 iterations;
and the drug protocol uses $\varepsilon=0.05$ and 100 iterations.
For genetics, the plan is computed from detached latents separately within
each record, and 128 pairs are sampled with replacement. Gradients pass
through the gathered latent values but not through the plan or sampled indices.
Random-pairing
controls independently sample the two endpoints, and no-OT controls remove the
transport plan while retaining the conditional flow architecture.

\subsection{Reconstruction and auxiliary losses}
\label{app:loss-details}

Endpoint reconstruction balances zero and nonzero expression entries. For
valid positions $\Omega$, let $\Omega_0$ and $\Omega_+$ denote entries with
zero and nonzero targets. With Huber threshold one, the loss for one endpoint
is
\begin{equation}
    \mathcal L_{\mathrm{rec,end}}
    =\frac{1}{2|\Omega_0|}
      \sum_{r\in\Omega_0}\ell_{\mathrm{Huber}}(\hat x_r,x_r)
    +\frac{1}{2|\Omega_+|}
      \sum_{r\in\Omega_+}\ell_{\mathrm{Huber}}(\hat x_r,x_r).
    \label{app:eq:reconstruction-loss}
\end{equation}
Source and target endpoints are reconstructed from their own latent states, so
this term does not create row-wise supervision across populations.

The genetic instance uses the complete objective
\begin{align}
    \mathcal L_{\mathrm{gene}}
    ={}&\mathcal L_{\mathrm{FM}}
    +\mathcal L_{\mathrm{rec,src}}+\mathcal L_{\mathrm{rec,tgt}}
    +100\mathcal L_{\mathrm{mean}}+\mathcal L_{\Delta\text{-corr}} \notag\\
    &+0.1\mathcal L_{\mathrm{flow\text{-}moment}}
    +0.1\mathcal L_{\mathrm{direction}}
    +0.05\mathcal L_{\mathrm{magnitude}} \notag\\
    &+0.1\mathcal L_{\mathrm{var}}
    +0.05\mathcal L_{\mathrm{proj\text{-}cov}}.
    \label{app:eq:gene-loss}
\end{align}
These coefficients are shared by all nine reported genetic training runs.
Here $\mathcal L_{\mathrm{mean}}$ is Smooth L1 loss between predicted and
observed population means in log-expression space, and
$\mathcal L_{\Delta\text{-corr}}$ is one minus Pearson correlation between
their mean shifts relative to the same source population. The flow-moment
term sums Smooth L1 losses on the means and log variances of the generated
and observed target latents. If $\hat d$ and $d$ are their latent mean
displacements from the source mean, the direction and magnitude terms are
$1-\cos(\hat d,d)$ and Smooth L1 loss between $\log\|\hat d\|_2$ and
$\log\|d\|_2$, respectively. The source reference and observed displacement
are detached for these two terms.

The variance loss compares log gene variances. The covariance loss compares
covariance matrices after a fixed $5041\times128$ Gaussian projection with
seed 42, scaling both matrices by the observed projected
standard-deviation products before applying Smooth L1. Variances and
normalization denominators use the implementation's $10^{-6}$ floor.
Smooth L1 uses threshold one and averages over valid entries; all population
statistics exclude padded cells. Reconstruction uses
Equation~\ref{app:eq:reconstruction-loss}, with an empty zero or nonzero
group contributing zero. Paired-expression supervision, separate latent
variance/covariance regularizers, population separation, and temporal
ordering are disabled for the genetic instance. The genetic results evaluate
this complete training recipe; the reported benchmark comparisons do not
isolate the contributions of OT pairing, continuous transport, or individual
auxiliary objectives.

\begin{table}[!htbp]
  \caption{Shared configuration for the nine genetic runs. Each holdout regime
  is trained independently with seeds 42, 123, and 2026.}
  \label{app:tab:genetic-config}
  \centering\small
  \begin{tabular}{p{0.43\linewidth}p{0.49\linewidth}}
    \toprule
    Component & Setting \\
    \midrule
    Expression encoder; cell latent & MLP $5041\rightarrow512\rightarrow256$ \\
    Decoder & MLP $256\rightarrow512\rightarrow5041$; Softplus \\
    Perturbation representation & Separate 512-dimensional gene embeddings; mean pooling \\
    Dropout & Encoder/decoder 0.1; vector field 0 \\
    OT & Median-normalized squared latent cost; $\varepsilon=0.05$; 30 iterations \\
    Batch & 8 conditions; 128 source and target cells per record \\
    Optimizer; maximum updates & AdamW, learning rate $10^{-4}$; 50,000 \\
    Checkpoint selection & Validation condition-mean L2; EMA decay 0.99 \\
    Inference & 8 Euler steps; no added initial noise; source-anchored decoding \\
    \bottomrule
  \end{tabular}
\end{table}

The drug instance uses
\begin{align}
    \mathcal L_{\mathrm{drug}}
    ={}& \mathcal L_{\mathrm{FM}}
    +\mathcal L_{\mathrm{mean}}
    +0.3\mathcal L_{\Delta\text{-corr}}
    +0.2\mathcal L_{\mathrm{var}}
    +0.05\mathcal L_{\mathrm{proj\text{-}cov}} \notag\\
    &+0.1\mathcal L_{\mathrm{flow\text{-}moment}}
    +0.05\mathcal L_{\mathrm{rec,src}}
    +0.05\mathcal L_{\mathrm{rec,tgt}}
    +0.1\mathcal L_{\mathrm{target\text{-}rec\text{-}moment}}.
    \label{app:eq:drug-loss}
\end{align}
The projected-covariance term uses a fixed $2000\times128$ random projection
with seed 42. These coefficients were selected using validation data before the
drug evaluation.

\subsection{Temporal ordering and intermediate requests}
\label{app:temporal-ordering}

The temporal flow-matching term uses Day~0 and Day~3 endpoint batches with the
descriptor $c_{\mathrm{end}}=(0,3,3)$. An auxiliary term evaluates virtual time
conditions without reading measured intermediate expression. Let $\bar z_0$
and $\bar z_3$ be endpoint population means and define
\begin{equation}
    d=\frac{\bar z_3-\bar z_0}
    {\lVert\bar z_3-\bar z_0\rVert_2}.
    \label{app:eq:trajectory-direction}
\end{equation}
For $t_k\in\{1,2,3\}$, the auxiliary branch substitutes
$(t_s,t_k,t_k-t_s)$ into the time fields and integrates the corresponding vector
field over the full interval $[0,1]$; the Day~3 prediction reuses the endpoint
rollout. It then mean-pools the predicted latent population
$\widehat{\bar z}_{t_k}$. Its progress score is
\begin{equation}
    s_k=(\widehat{\bar z}_{t_k}-\bar z_0)^\top d,
    \label{app:eq:trajectory-score}
\end{equation}
and adjacent scores are separated by margin $m=0.02$:
\begin{equation}
    \mathcal L_{\mathrm{traj}}
    =\frac{1}{2}\sum_{k=1}^{2}
      \max\!\left(0,m-(s_{k+1}-s_k)\right).
    \label{app:eq:trajectory-loss}
\end{equation}
The full temporal flow objective is
\begin{equation}
    \mathcal L_{\mathrm{time}}
    =\mathcal L_{\mathrm{FM}}
    +0.1\mathcal L_{\mathrm{rec,src}}
    +0.1\mathcal L_{\mathrm{rec,tgt}}
    +0.1\mathcal L_{\mathrm{traj}}.
    \label{app:eq:time-loss}
\end{equation}

The static OT residual comparator retains OT endpoint sampling and the
transition MLP dimensions. It evaluates the transition network at the source
latent with its scalar time input fixed to one, defining
$r_\theta(z_0,g_s,\eta)=v_\theta(z_0,1,g_s,\eta)$, and regresses this output
against $z_1-z_0$. A requested horizon $h$ gives the single update
$\widehat z_h=z_0+h\,r_\theta(z_0,g_s,\eta)$. It retains endpoint
reconstruction but does not use interpolated-state velocity regression or the
flow-specific ordering term. Thus, this comparator changes the training
formulation as well as the generation rule; it is not an integrator-only
ablation. The random-pairing flow variant retains the flow formulation and
substitutes independently paired endpoints for OT samples.

Intermediate prediction keeps the endpoint descriptor fixed and maps request
time to an integration horizon:
\begin{equation}
    r(t_r)=\frac{t_r-t_s}{t_e-t_s},
    \qquad
    h(t_r)=r(t_r)^\gamma,
    \qquad \gamma=0.5.
    \label{app:eq:temporal-horizon}
\end{equation}
For the Day~0--Day~3 protocol, $h(1)=0.577$, $h(2)=0.816$, and $h(3)=1$.
The endpoint-conditioned field is integrated over $[0,h(t_r)]$ with a
32-step Heun solver. The square-root warp is fixed by the task protocol and is
not learned biological time.

Representation pretraining and transition-module training both use only
Day~0 and Day~3 populations from training donors, as specified in the main
temporal protocol. Measured Day~1 and Day~2 expression is excluded from both
training stages and from transition checkpoint selection. Transition
checkpoints are selected by raw RBF MMD on validation
Day~0$\rightarrow$Day~3 endpoint predictions. At inference, each held-out
donor supplies only Day~0 expression and the requested time; measured
intermediate populations are used only after generation for evaluation.
Section~\ref{sec:temporal-sensitivity} reports interventions that exchange or
fix the requested horizons.

\subsection{Task-specific inference}
\label{app:task-inference-details}

\begin{table}[!htbp]
    \centering
    \caption{Task-specific representation and inference settings.}
    \label{app:tab:task-inference}
    \small
    \setlength{\tabcolsep}{5pt}
    \renewcommand{\arraystretch}{1.12}
    \begin{tabular}{@{}lrrlll@{}}
        \toprule
        Task & Genes & Latent & Source encoder & Solver & Steps \\
        \midrule
        Temporal & 2,009 & 256 & Trainable & Heun & 32 \\
        Genetic  & 5,041 & 256 & Trainable & Euler & 8 \\
        Drug     & 2,000 & 512 & Frozen in stage B & Heun & 16 \\
        \bottomrule
    \end{tabular}
\end{table}
Temporal requests use partial integration as defined above. The genetic
model takes and returns expression in the same ordered 5,041-gene space.
Mean L2, perturbation-effect correlation, and Energy are evaluated on this
full model gene set. DE $\rho$ is evaluated on condition-specific
reference DE genes within this space.
The 5,041 genes comprise 5,000 highly variable genes selected from the full
quality-controlled dataset before condition splitting and 41 additional
measured perturbation targets. The drug model is trained in
two stages: expression reconstruction followed by conditional transition
learning. Stage B freezes the source encoder and gene embedding, trains the
descriptor encoder, velocity field, and decoder, and uses source-anchored
residual decoding. The descriptor encoder is warm-started by response-aware
pretraining on training-population responses, with validation used for selection.
Only its condition modules are transferred; the temporary response head is
discarded. This initialization does not make held-out target expression an
inference input. Transition learning continues to update the descriptor encoder.
The transition solver adds standard-normal cell-level latent noise, centered
over valid source cells so that their latent mean is unchanged. The same
initial-noise rule applies to auxiliary generated rollouts during training
and to inference. OT pairing and the FM straight-line paths use the encoded
endpoints before this rollout noise is added. In every task, target expression from the held-out unit is absent
from the inference inputs.

\subsection{Drug adapter and evaluation interfaces}
\label{app:drug-adapter-interface}

The chemCPA adapter~\citep{hetzel2022chemcpa} uses ComPert with fixed 256-dimensional SMILES-derived
features, an amortized doser, and cell-line covariates. Predictions are sampled
from the predicted Gaussian expression distribution. Treatment
duration is not an explicit input to this adapter. The scDFM adapter~\citep{yu2026scdfm} uses a
continuous molecular-feature, dose, duration, and cell-line interface around
the PAD-Transformer backbone. The CellFlow adapter~\citep{klein2025cellflow} uses the same types of
condition inputs with an OT flow-matching model and a 50-component PCA
representation fitted on training cells. All three are project adaptations,
not unmodified reference benchmarks.

The common evaluator receives predictions on the ordered 2,000-gene
expression axis. Its distribution distances use a separate Gaussian random
projection with 50 columns and seed 42; this is not CellFlow's internal PCA
space. The multiscale RBF bandwidth is calibrated on training populations.
This evaluation projection is also distinct from the 128-dimensional
projection used by the drug model's covariance training loss. Adapter-specific
generation procedures are retained; outputs are not forced to share a common
sampling model.

\subsection{Additional drug-response analyses}
\label{app:drug-evidence}
\label{app:drug-seed-results}

The supplementary Stage-B analysis uses training seeds 42, 123, and 2026
with a shared frozen seed-42 expression backbone and condition warm-start.
Table~\ref{tab:drug-seed-frozen-primary} reports condition-weighted scores
for each trained checkpoint. The aggregate pMMD and pEnergy are
$0.031\pm0.001$ and $0.197\pm0.010$; these SD describe training-seed
variation, separately from the fixed-model test repeats in the main comparison.

\begin{table}[!htbp]
  \caption{Seed-level Stage-B results for \textsc{TempoBridge} on the
  compound-disjoint sci-Plex3 split, using a shared frozen seed-42 expression
  backbone and condition warm-start. Checkpoints were selected on validation
  data before evaluation on the compound-disjoint split. Each seed's score
  weights conditions equally; SD is the sample SD across the three training
  seeds, not a compound-bootstrap interval.}
  \label{tab:drug-seed-frozen-primary}
  \centering
  \scriptsize
  \setlength{\tabcolsep}{2.8pt}
  \begin{tabular}{@{}lrrrrr@{}}
    \toprule
    Metric & Seed 42 & Seed 123 & Seed 2026 & Mean & SD \\
    \midrule
    L2$\downarrow$       & 4.515 & 4.498 & 4.513 & 4.509 & 0.010 \\
    MSE$\downarrow$      & 0.011 & 0.011 & 0.011 & 0.011 & $<0.001$ \\
    MAE$\downarrow$      & 0.074 & 0.074 & 0.075 & 0.074 & $<0.001$ \\
    Gene $r\uparrow$     & 0.976 & 0.976 & 0.977 & 0.976 & $<0.001$ \\
    $\Delta r\uparrow$   & 0.418 & 0.417 & 0.441 & 0.425 & 0.014 \\
    DE $\rho\uparrow$    & 0.340 & 0.285 & 0.277 & 0.301 & 0.034 \\
    DE@100$\uparrow$     & 0.261 & 0.257 & 0.282 & 0.267 & 0.013 \\
    \bottomrule
  \end{tabular}
\end{table}

The main comparison evaluates the full two-stage drug pipeline and uses one
fixed checkpoint per method with
test seeds 42, 123, and 2026. Its condition-weighted pEnergy means are
0.2123 for TempoBridge, 0.2260 for Conditional MLP, and 0.2219 for CFM
without OT (Table~\ref{tab:drug-main}). The separate paired analysis first
averages conditions within each compound, then weights the 19 compounds
equally. Its pEnergy intervals against both baselines cross zero. These
compound-weighted intervals concern a different average from the
condition-weighted table means, so the main comparison supports a mean
ranking without establishing a compound-level advantage.
Appendix~\ref{app:uncertainty-reporting} specifies the bootstrap; it holds
the fitted models fixed and does not represent training-seed variability.

\begin{table}[!htbp]
  \caption{Drug-response main evaluation and supplementary Stage-B analysis.
  All rows weight biological conditions equally. The Stage-B row averages
  transition modules fitted with training seeds 42, 123, and 2026 on a shared
  frozen seed-42 expression backbone and condition warm-start. The
  main-pipeline and baseline rows evaluate one fixed checkpoint per method
  with test seeds 42, 123, and 2026, as in Table~\ref{tab:drug-main}; the
  main-pipeline row additionally reports variance and covariance correlations.
  Their SDs describe training-seed and test-run variation, respectively.
  Cross-row differences do not isolate an OT component effect.}
  \label{tab:drug-training-stability}
  \centering
  \scriptsize
  \setlength{\tabcolsep}{2.6pt}
  \resizebox{\linewidth}{!}{%
  \begin{tabular}{@{}lrrrrrrr@{}}
    \toprule
    Method & pEnergy$\downarrow$ & pMMD$\downarrow$ & L2$\downarrow$ & $\Delta r\uparrow$ & DE@100$\uparrow$ & Var. $r\uparrow$ & Cov. $r\uparrow$ \\
    \midrule
    Stage-B TempoBridge & 0.197$\pm$0.010 & 0.031$\pm$0.001 & 4.509$\pm$0.010 & 0.425$\pm$0.014 & 0.267$\pm$0.013 & 0.909$\pm$0.003 & 0.219$\pm$0.002 \\
    Main-pipeline TempoBridge & 0.212$\pm$0.019 & 0.033$\pm$0.004 & 4.248$\pm$0.047 & 0.598$\pm$0.005 & 0.547$\pm$0.013 & 0.908$\pm$0.004 & 0.221$\pm$0.001 \\
    Conditional MLP & 0.226$\pm$0.010 & 0.032$\pm$0.001 & 4.878$\pm$0.046 & 0.225$\pm$0.019 & 0.322$\pm$0.023 & 0.937$\pm$0.000 & 0.343$\pm$0.001 \\
    Conditional CFM (no OT) & 0.222$\pm$0.009 & 0.034$\pm$0.003 & 4.636$\pm$0.029 & 0.385$\pm$0.006 & 0.263$\pm$0.006 & 0.904$\pm$0.002 & 0.217$\pm$0.019 \\
    \bottomrule
  \end{tabular}%
  }
\end{table}

The matched $2\times2$ conditioning analysis varies source context and
cell-line information within the fixed-backbone Stage-B setting. Its three
training seeds share the frozen seed-42 expression backbone and response-aware
condition initialization; the descriptor encoder, field, and decoder are then
trained for each seed. Table~\ref{tab:drug-context-ablation} separates
condition-weighted means and training-seed SD from compound-weighted paired
effects. Full-minus-No-context differences were 0.004 for $\Delta r$
(95\% CI $[0.001,0.006]$) and 0.019 for DE $\rho$ ($[0.008,0.033]$).
Under condition weights, pEnergy favored No source context and pMMD was
unchanged at the reported precision; removing cell-line information reduced
$\Delta r$ from 0.425 to 0.414. The compound-weighted effects and
condition-weighted changes summarize different averages.

\begin{table}[!htbp]
  \caption{Matched $2\times2$ conditioning ablation on compound-disjoint
  sci-Plex3 in the fixed-backbone Stage-B setting. (a) Equal condition weights
  within each training seed; mean$\pm$sample SD over seeds 42, 123, and 2026.
  (b) Compound-weighted paired differences, Full minus No source context,
  with their 95\% intervals; positive values favor Full. These differences
  use compound weights and are not obtained by subtracting the displayed
  condition-weighted means. Removing a condition replaces its vector with
  zeros without changing model capacity.}
  \label{tab:drug-context-ablation}
  \centering
  \scriptsize
  \setlength{\tabcolsep}{3.0pt}
  \textbf{a) Condition-weighted means and training-seed SD}\par\smallskip
  \begin{tabular}{@{}lrrrrrr@{}}
    \toprule
    Arm & pEnergy$\downarrow$ & pMMD$\downarrow$ & L2$\downarrow$ & $\Delta r\uparrow$ & DE $\rho\uparrow$ & DE@100$\uparrow$ \\
    \midrule
    Full & 0.197$\pm$0.010 & 0.031$\pm$0.001 & 4.509$\pm$0.010 & \textbf{0.425$\pm$0.014} & \textbf{0.301$\pm$0.034} & 0.267$\pm$0.013 \\
    No source context & \textbf{0.193$\pm$0.005} & $0.031\,(\mathrm{SD}<0.001)$ & \textbf{4.506$\pm$0.017} & 0.422$\pm$0.004 & 0.289$\pm$0.046 & 0.261$\pm$0.003 \\
    No cell line & 0.195$\pm$0.003 & $0.031\,(\mathrm{SD}<0.001)$ & 4.539$\pm$0.006 & 0.414$\pm$0.003 & 0.295$\pm$0.055 & 0.260$\pm$0.005 \\
    Neither & 0.196$\pm$0.004 & $0.031\,(\mathrm{SD}<0.001)$ & 4.538$\pm$0.008 & 0.415$\pm$0.004 & 0.295$\pm$0.052 & \textbf{0.270$\pm$0.003} \\
    \bottomrule
  \end{tabular}
  \par\medskip
  \textbf{b) Compound-weighted paired effects: Full minus No source context}\par\smallskip
  \begin{tabular}{@{}lcc@{}}
    \toprule
    Metric & Paired mean difference & 95\% interval \\
    \midrule
    $\Delta r$ & 0.004 & $[0.001,0.006]$ \\
    DE $\rho$ & 0.019 & $[0.008,0.033]$ \\
    Condition-PDS & 0.005 & $[0.002,0.008]$ \\
    \bottomrule
  \end{tabular}
\end{table}

Condition-PDS evaluates response identification among the full set of
biological conditions. For each predicted mean response, candidates are the
observed condition-specific mean responses relative to their source means,
ranked by mean absolute (L1) distance. With $K$ candidates and correct-condition
rank $r$, the score is $1-(r-1)/(K-1)$; average ranks handle ties, and larger
scores are better. For the Stage-B paired analysis, condition-level metric
scores are first averaged across the fitted training-seed models. Full-minus-
No-context differences are then averaged within compound. A percentile
bootstrap resamples whole compounds 10,000 times (seed 20260824), holding
the fitted models fixed. These intervals describe compound variation after
seed averaging and are distinct from the training-seed SD in panel (a).

The fixed-backbone projection analysis evaluates Gaussian dimensions 20, 50,
100, and 200 with five train-only projection seeds each, followed by PCA-50 and
PCA-100. TempoBridge has lower Energy than Conditional MLP in all 20 Gaussian
spaces and 21 of the 22 spaces overall. Under PCA-50, TempoBridge's Energy
and MMD are approximately $2.74\%$ and $7.23\%$ higher than MLP's,
respectively. Under PCA-100, its Energy is approximately $1.41\%$ lower
and its MMD $3.36\%$ higher. MMD favors TempoBridge in 8 of 22 spaces.
The magnitude and direction of the distributional comparisons therefore depend
on the evaluation projection.

\begin{table}[!htbp]
  \caption{Fixed-backbone TempoBridge projection-space sensitivity relative to
  Conditional MLP. Each Gaussian dimension uses five train-only projection
  seeds. The final row also includes PCA-50 and PCA-100.}
  \label{tab:drug-projection-robustness}
  \centering
  \small
  \begin{tabular}{@{}lrrr@{}}
    \toprule
    Evaluation space & Energy win rate & Mean Energy reduction & MMD win rate \\
    \midrule
    Gaussian 20D & 100.000\% & 7.530\% & 20.000\% \\
    Gaussian 50D & 100.000\% & 8.280\% & 40.000\% \\
    Gaussian 100D & 100.000\% & 9.640\% & 60.000\% \\
    Gaussian 200D & 100.000\% & 10.800\% & 40.000\% \\
    All 22 spaces & 95.500\% (21/22) & 8.180\% & 36.400\% (8/22) \\
    \bottomrule
  \end{tabular}
\end{table}

Finally, a scaffold-disjoint split tests stricter chemical extrapolation.
This supplementary analysis uses seed 42 only.

\begin{table}[!htbp]
  \caption{Supplementary scaffold-disjoint evaluation for seed 42. The test
  split contains 17 compounds and 220 biological conditions. This single-seed
  analysis is not used for the main compound-disjoint claim.}
  \label{tab:drug-scaffold}
  \centering
  \scriptsize
  \setlength{\tabcolsep}{3.5pt}
  \begin{tabular}{@{}lrrrrr@{}}
    \toprule
    Method & L2$\downarrow$ & $\Delta r\uparrow$ & DE $\rho\uparrow$ & pMMD$\downarrow$ & pEnergy$\downarrow$ \\
    \midrule
    TempoBridge & \textbf{4.677} & \textbf{0.414} & 0.161 & 0.034 & \textbf{0.223} \\
    Conditional CFM (no OT) & 4.793 & 0.387 & 0.171 & 0.035 & 0.242 \\
    Conditional MLP & 5.060 & 0.239 & 0.361 & 0.034 & 0.259 \\
    Fingerprint KNN & 5.187 & 0.225 & \textbf{0.388} & 0.037 & 0.279 \\
    chemCPA adapter & 5.620 & 0.368 & 0.386 & 0.046 & 0.386 \\
    \bottomrule
  \end{tabular}
\end{table}

\section{Temporal Differentiation Details}
\label{app:temporal-details}

\subsection{Dataset and preprocessing}
\label{app:data}

The data split contained 21,533 training cells from 77 donors, 3,556
validation cells from 17 donors, and 10,955 cells from 32 donors in the
evaluation partition. Requiring at least 64 cells at each
of Days 0, 1, and 2 yielded the fixed 15-donor evaluation cohort used by every
method. Cells were retained if they had at least 500 total counts, 200 detected
genes, and a mitochondrial fraction below 0.2; genes had to occur in at least
three cells. Counts were
normalized to a library size of 10,000 and transformed with $\log(1+x)$. Highly
variable genes were selected without using the evaluation cohort. The nine
prespecified genes were added to the 2,000-gene set before model training.

\subsection{Architecture and optimization}
\label{app:model-details}

The expression encoder has widths
$2009\!\rightarrow\!1024\!\rightarrow\!256$; the decoder mirrors this mapping
with a Softplus output. The velocity MLP uses two hidden transformations after a
769-dimensional concatenated input. The complete model contains 9,797,593
parameters. Three-seed training used weights 1.0, 0.1, 0.1, and 0.1 for
flow matching, source reconstruction, target reconstruction, and trajectory
consistency, respectively. The trajectory term evaluated virtual target times
$\{1,2,3\}$ with a ranking margin of 0.02; it did not use measured Day~1 or
Day~2 expression. Optimization used AdamW with a base learning rate of
$2\times10^{-4}$, a learning rate of $2\times10^{-5}$ for pretrained
parameters, weight decay 0.005, gradient clipping at 1.0, a batch size of 256,
and bfloat16 arithmetic. Mean training cost was $2.467\pm0.896$ GPU-hours.
With 32-step Heun integration, inference took $0.032\pm0.001$ seconds per
population or $0.374\pm0.170$ seconds per 1,000 cells in the timing experiment.

Checkpoint selection used validation Day~0$\rightarrow$Day~3 endpoints. Models
were evaluated every 500 steps and selected by raw single-bandwidth RBF MMD,
with patience 8 after two warm-up evaluations and a minimum improvement of
0.001. The selected steps for seeds 42, 123, and 2026 were 8,500, 10,500, and
4,000. These records identify independently trained checkpoint candidates;
the main comparison uses one selected checkpoint for all three test repeats.
Main-table
uncertainty follows the fixed-model randomized-test-run definition in
Section~\ref{sec:experimental-setup}, rather than treating checkpoint
identifiers as test-repeat identifiers.

\subsection{Representation and gene-space sensitivity}
\label{app:representation}

The temporal task was rescored on the full 2,009-gene training space,
the subset of 2,000 highly variable genes, and the nine-gene prespecified
panel. This analysis checks whether the conclusion depends on one favorable
feature set. A scratch-encoder control was also trained for three seeds. The
pretrained initialization did not produce a stable advantage and is not
presented as a contribution.

\subsection{Source-context ablation protocol}
\label{app:source-context-ablation}

The temporal source-context ablation uses the same 2,009-gene space, 15-donor
evaluation cohort, and 30 Day~0$\rightarrow$Day~1/2 requests as the main
comparison. This separate retraining diagnostic uses training seeds 42, 123,
and 2026, with both arms trained independently of the main-table models.
Within each training seed, the Full and No-context arms use the same
training configuration except for source-context conditioning.
The No-context arm replaces the
pooled vector $h(\widehat P_s)$ with a zero vector while retaining source-cell
latents, the expression encoder, transition descriptor, OT construction, flow
objective, decoder, seed, and training protocol. Aggregate descriptive results
are reported in Table~\ref{tab:source-context-ablation-2009}.

The matched Full arm has Formal MMD $0.378\pm0.021$ and Energy
$0.135\pm0.005$, compared with the independently trained main-table
Full Flow values $0.363\pm0.014$ and
$0.129\pm0.005$. The context-free arm has larger between-seed
standard deviations for Formal MMD ($0.048$ versus $0.021$), Energy
($0.013$ versus $0.005$), and variance correlation ($0.047$ versus $0.010$).
The ablation compares the two matched arms descriptively; the main-table
Full Flow checkpoint is not part of this matched comparison. The supplementary
between-training-seed standard deviations and the main-table test-run
standard deviations describe different sources of variability and should
not be compared as if they estimated the same uncertainty.

\begin{table}[!htbp]
  \caption{Separately matched source-context ablation on the 15-donor temporal
  cohort in the 2,009-gene space. Both arms were retrained under the same
  training configuration and seeds, independently of the main-table models;
  the only difference is whether the pooled source context is supplied. Values
  are mean $\pm$ sample standard deviation over three training seeds. Relative change
  is signed so that positive values favor the full-context arm.}
  \label{tab:source-context-ablation-2009}
  \centering
  \small
  \setlength{\tabcolsep}{5.0pt}
  \begin{tabular}{@{}lrrr@{}}
    \toprule
    Metric & Full context & No context & Relative change \\
    \midrule
    Formal MMD $\downarrow$ & \textbf{0.378$\pm$0.021} & 0.397$\pm$0.048 & \textbf{4.720\%} \\
    Energy $\downarrow$ & \textbf{0.135$\pm$0.005} & 0.139$\pm$0.013 & \textbf{3.100\%} \\
    $\Delta$ Pearson $\uparrow$ & 0.625$\pm$0.005 & \textbf{0.639$\pm$0.010} & -2.170\% \\
    Variance Pearson $\uparrow$ & \textbf{0.622$\pm$0.010} & 0.613$\pm$0.047 & \textbf{1.620\%} \\
    Covariance Pearson $\uparrow$ & 0.419$\pm$0.006 & \textbf{0.421$\pm$0.005} & -0.470\% \\
    DE@100 $\uparrow$ & 0.515$\pm$0.005 & \textbf{0.529$\pm$0.007} & -2.730\% \\
    \bottomrule
  \end{tabular}
\end{table}

\subsection{Temporal donor context-swap intervention}
\label{app:temporal-context-swap}

The matched retraining experiment asks whether the context channel has a net
effect. A separate inference-time intervention asks whether a frozen model uses
the context of the requested donor. For each of the 15 donors, Day~0 cells were
split into nonoverlapping query and context subsets. Each Day~1 and Day~2
request was evaluated with the correct donor context and with contexts from
each of the other 14 donors in each of three seeded evaluations (42, 123,
and 2026). Within each donor/day/seed request, the same checkpoint, query
cells, and targets are used for all context choices. The checkpoint is held
fixed throughout each paired intervention. This gives $15\times2\times15\times3=1{,}350$
evaluated inference cases. No parameter was updated and no donor was excluded
after scoring.

\begin{table}[!htbp]
  \caption{Temporal donor context-swap intervention. The correct context is
  compared with the mean over the 14 incorrect donor contexts while query
  cells, requested time, checkpoint, decoder, and target remain fixed within
  each donor/day/seed comparison. Effects average the two days and evaluations
  with seeds 42, 123, and 2026, and are reported in the direction favoring the
  correct context; intervals use donor-level paired bootstrap conditional on
  the fitted checkpoints and context bank.}
  \label{tab:temporal-context-swap}
  \centering
  \scriptsize
  \setlength{\tabcolsep}{3.2pt}
  \begin{tabular}{@{}lrrrrr@{}}
    \toprule
    Metric & Correct & Incorrect mean & Effect & 95\% CI & Donors won \\
    \midrule
    Energy$\downarrow$ & 0.138 & 0.142 & 0.004 & [0.002,0.007] & 12/15 \\
    Formal MMD$\downarrow$ & 0.385 & 0.392 & 0.007 & [0.004,0.009] & 12/15 \\
    $\Delta r\uparrow$ & 0.627 & 0.616 & 0.011 & [0.003,0.021] & 11/15 \\
    DE@100$\uparrow$ & 0.512 & 0.503 & 0.009 & [0.004,0.014] & 12/15 \\
    \bottomrule
  \end{tabular}
\end{table}

The correct context reduced Energy by $0.00292$ at Day~1 (95\% CI
$[0.00080,0.00511]$) and by $0.00556$ at Day~2 (95\% CI
$[0.00259,0.00871]$). Both effects and intervals are absolute Energy
differences. The correct-context retrieval rate was
17.780\% at rank one and 45.560\% within the top three, compared with a 6.670\%
random rank-one rate. These diagnostics indicate use of
donor-specific population information; they do not establish donor identity
classification or a causal biological effect.

\subsection{Metric definitions and statistical unit}
\label{app:metrics}

For genetic evaluation, Control copies the first $n$ cells of the fixed
128-cell source panel, with $n$ matching the target cardinality. Effect metrics
use the full 128-cell source mean as their reference. Hence Control has a
zero mean effect, and undefined effect correlation, when $n=128$; smaller
target populations can yield finite correlations from the prefix/reference
difference. Summary correlations exclude undefined conditions. These values
do not represent learned perturbation responses.

The following definitions distinguish the three real-data evaluators.
Let $X=\{x_i\}_{i=1}^{n}$ and $Y=\{y_j\}_{j=1}^{m}$ denote generated and
observed target cells, and let $S$ be the observed source population.
Mean L2 is $\|\bar x-\bar y\|_2$ in the declared log-expression space;
$\Delta r$ is Pearson correlation across genes between
$\bar x-\bar s$ and $\bar y-\bar s$. The temporal, genetic, and drug
main comparisons use 2,009, 5,041, and 2,000 genes, respectively.
Metrics are computed within a biological condition before aggregation.

\paragraph{Energy distance.}
For representations $a_i=T(x_i)$ and $b_j=T(y_j)$, we estimate Energy distance~\citep{szekely2013energy} with the following U-statistic:
\begin{equation}
 \widehat{\mathcal E}_{U}
 =\frac{2}{nm}\sum_{i,j}\|a_i-b_j\|_2
 -\frac{1}{n(n-1)}\sum_{i\ne i'}\|a_i-a_{i'}\|_2
 -\frac{1}{m(m-1)}\sum_{j\ne j'}\|b_j-b_{j'}\|_2.
 \label{app:eq:energy-estimator}
\end{equation}
The temporal evaluator uses $T(x)=x/\sqrt{G_{\mathrm{eval}}}$, where
$G_{\mathrm{eval}}$ is the number of scored genes (2,009 for the full-space
comparisons), and the V-statistic:
the within-population sums include diagonal entries and have denominators
$n^2$ and $m^2$. Its final estimate is clipped at zero. The genetic
evaluator uses $T(x)=x$ and Equation~\ref{app:eq:energy-estimator}, retaining
potentially negative finite-sample estimates. Drug projected Energy uses
$T(x)=xR$, where $R\in\mathbb R^{2000\times50}$ has independent
$\mathcal N(0,1/50)$ entries generated with seed 42. It uses the U-statistic
followed by a nonnegative clamp. None of these Energy scores takes an outer
square root. Their absolute magnitudes are not comparable across tasks or
gene panels.

\paragraph{MMD, bandwidths, and clipping.}
With $k_\sigma(a,b)=\exp[-\|a-b\|_2^2/(2\sigma^2)]$, the U-statistic estimator of squared MMD~\citep{gretton2012kernel} is
\begin{equation}
 \widehat M^2_U(\sigma)
 =\frac{\sum_{i\ne i'}k_\sigma(a_i,a_{i'})}{n(n-1)}
 +\frac{\sum_{j\ne j'}k_\sigma(b_j,b_{j'})}{m(m-1)}
 -\frac{2\sum_{i,j}k_\sigma(a_i,b_j)}{nm}.
 \label{app:eq:mmd-estimator}
\end{equation}
Temporal Formal MMD uses the same scaled expression space as temporal
Energy and bandwidth factors $\{0.25,0.5,1,2,4\}$. Each population is
capped at 512 cells by lexicographically sorting expression rows and
taking evenly spaced indices. The base bandwidth is the square root of
the median positive squared distance among pooled within- and
cross-population pairs. Kernels are accumulated in blocks of 256 cells.
The reported value is
$\sqrt{\frac{1}{5}\sum_a\max\{0,\widehat M^2_U(a\sigma_0)\}}$:
the implementation clips each bandwidth's estimate before averaging.
Raw \texttt{mmd\_rbf}, used for temporal endpoint checkpoint selection,
instead takes the square root of a nonnegative-clipped, single-bandwidth
V-statistic. The main-table Formal MMD field is
\texttt{formal\_mmd\_rbf\_multiscale}; it is distinct from that selection
metric.

Genetic MMD uses unscaled log expression and factors $\{0.5,1,2,4\}$.
For each regime, the base bandwidth is fixed from 2,048 pooled validation
target cells before test evaluation; the base values for Seen 2/2,
Seen 1/2, and Seen 0/2 are 17.76577, 16.78579, and 18.10347, respectively.
Drug pMMD uses the fixed Gaussian-50 projection and the same four factors,
with its base bandwidth calibrated on training cells. These two evaluators
report $\sqrt{\max\{0,\frac{1}{4}\sum_a\widehat M^2_U(a\sigma_0)\}}$,
clipping after averaging over bandwidths. The 128-dimensional projection
used by the covariance training loss is separate from the drug evaluation
projection. A quantity explicitly labeled MMD$^2$ has no outer square root.

\paragraph{Differential-expression recovery.}
Temporal DE@100 selects the 100 largest absolute changes in the mean
log-expression vector, separately for prediction and observation relative
to the same Day~0 mean, using \texttt{torch.topk}. It reports the
intersection size divided by 100; it does not filter genes by a
significance test. Drug DE@100 uses the same overlap denominator but ranks
absolute pseudobulk log$_2$ fold changes. Ties use stable sorting in the
fixed gene order. The genetic evaluator's auxiliary
\texttt{de\_overlap\_100} field instead reports the Jaccard index of its
top-100 fold-change sets; it is not the temporal DE@100 metric and is not
the DE $\rho$ reported in Tables~\ref{tab:gene-by-setting-full} and~\ref{tab:gene-main}.

For genetic and drug DE $\rho$, the fold change of population $X$ is
\begin{equation}
 \operatorname{LFC}_g(X,S)
 =\log_2\frac{\frac{1}{n}\sum_i\max\{\operatorname{expm1}(x_{ig}),0\}+\eta}
 {\frac{1}{|S|}\sum_{s\in S}\max\{\operatorname{expm1}(s_g),0\}+\eta}.
 \label{app:eq:lfc}
\end{equation}
The pseudocount $\eta$ is $10^{-8}$ for genetics and float64 machine
epsilon for drugs. Observed source and target cells alone determine the
DE gene set: genetics uses two-sided Mann--Whitney U tests with asymptotic
p-values and continuity correction, whereas drugs use Welch t-tests on
log-expression values. Both apply Benjamini--Hochberg correction with
$q<0.05$. DE $\rho$ is Spearman correlation of predicted and observed LFCs
on that set, using average ranks for ties and requiring at least 10 genes.
Undefined correlations remain missing and are excluded from the valid
metric denominator, rather than replaced by zero.

\paragraph{Aggregation and uncertainty.}
\label{app:uncertainty-reporting}
In the temporal main and daywise tables, each randomized test run first
averages the 15 donor-specific scores at each requested day, then averages
Day~1 and Day~2 equally. Genetic evaluation averages equally across retained
perturbation conditions within each setting and test run.
Table~\ref{tab:gene-by-setting-main} reports setting-specific means;
Table~\ref{tab:gene-main} additionally
reports the equally weighted average of the four settings. Drug evaluation
pools replicates within each cell-line/compound/dose/time condition before
scoring and averages the 248 conditions equally within each test run.
For each learned method and training setting, the same selected checkpoint
is used in all three test runs, with random seeds 42, 123, and 2026.
These seeds control the randomized evaluations, not three independently
trained models. Let $a_r$ denote the aggregated metric from
run $r\in\{1,2,3\}$. The reported mean is
$\bar a=(a_1+a_2+a_3)/3$, and its sample SD is
$s=\sqrt{\sum_{r=1}^{3}(a_r-\bar a)^2/(3-1)}$.
The SD is computed from run-level aggregates, rather than pooling cells,
conditions, or checkpoints across runs. It quantifies repeated-evaluation
variability conditional on the selected checkpoint; it is neither a standard
error nor a biological-unit confidence interval. The genetic tables report means without SD columns.
Separately labeled supplementary training-seed analyses aggregate one score
per independently trained checkpoint using the training-seed denominator
stated in the corresponding caption. Those checkpoints do not contribute
to the main-table test-run SD. Cells are not independent replicates for
these uncertainty summaries.

For context-swap intervals, the 14 incorrect contexts are first averaged
within each donor/day/seed request. Their paired difference from the
correct context is averaged across the three seeded evaluations within a donor; the overall
analysis additionally averages the two days. The 95\% percentile interval
uses 10,000 bootstrap resamples of the 15 donors, with seed 20260825.
For the drug main-pipeline comparisons, condition scores averaged across the
three test runs are first averaged within each compound, then 19 whole compounds are
resampled 10,000 times, with seed 20260825. Thus those intervals weight
compounds equally, whereas the drug main table weights conditions equally.
They estimate compound-weighted effects and are not confidence intervals
for the condition-weighted table means.
These intervals describe biological-unit variation conditional on the fitted models;
they do not resample training seeds or individual cells.

\subsection{Auxiliary time-adapted scVIDR reference}
\label{app:scvidr-reference}

\begin{table}[!htbp]
  \caption{Auxiliary 2,009-gene result for the time-adapted scVIDR
  implementation on the fixed 15-donor cohort with target-matched cardinality.
  Values follow the main temporal test-repeat protocol: mean $\pm$ sample
  standard deviation over test seeds 42, 123, and 2026.}
  \label{tab:scvidr-reference}
  \centering
  \small
  \setlength{\tabcolsep}{4pt}
  \begin{tabular}{lcccc}
    \toprule
    Method & Energy $\downarrow$ & Formal MMD $\downarrow$ & Sink. $\downarrow$ & Gene $r$ $\uparrow$ \\
    \midrule
    scVIDR (time-adapted) & $0.294\pm0.005$ & $0.476\pm0.006$ & $0.195\pm0.002$ & $0.726\pm0.005$ \\
    \bottomrule
  \end{tabular}

  \vspace{1mm}
  \begin{tabular}{lcccc}
    \toprule
    Method & $\Delta r$ $\uparrow$ & Var. $r$ $\uparrow$ & Cov. $r$ $\uparrow$ & DE@100 $\uparrow$ \\
    \midrule
    scVIDR (time-adapted) & $0.604\pm0.006$ & $0.546\pm0.015$ & $0.333\pm0.017$ & $0.694\pm0.003$ \\
    \bottomrule
  \end{tabular}
\end{table}

The scVIDR implementation adapts a perturbation-response model~\citep{kana2023scvidr} to the temporal
endpoint protocol and is reported separately from the primary
external-method ranking. The row uses the fixed 15-donor cohort, the
2,009-gene space, target-cardinality matching, and the three test repeats.

\subsection{Non-deployable endpoint references}
\label{app:oracle-references}

\begin{table}[!htbp]
  \caption{Non-deployable endpoint references. Both methods read held-out Day~3
  expression and are excluded from the external-method ranking.}
  \label{tab:oracle-references}
  \centering
  \small
  \begin{tabular}{lccccc}
    \toprule
    Method & Energy $\downarrow$ & Formal MMD $\downarrow$ & Gene $r$ $\uparrow$ & Var. $r$ $\uparrow$ & DE@100 $\uparrow$ \\
    \midrule
    Endpoint interpolation & 0.225 & 0.364 & 0.755 & 0.632 & 0.699 \\
    PCA interpolation & 0.290 & 0.466 & 0.724 & 0.598 & 0.688 \\
    \bottomrule
  \end{tabular}
\end{table}

Both interpolation references use held-out Day~3 expression and do not satisfy
the Day~0-only inference protocol. They are not included in the external-method
ranking. Their values provide endpoint-informed reference
points, not universal upper bounds: an interpolation rule can be favorable for
this four-day schedule without representing the correct transition mechanism.

\subsection{Supplementary temporal analyses}
\label{app:supplementary-temporal}

\begin{table}[!htbp]
  \caption{Request-time decomposition in the 2,009-gene space on the fixed
  15-donor cohort. Within each randomized test run, metrics are averaged
  equally across the 15 donor-specific requests at each day. Values are mean
    $\pm$ sample standard deviation over three randomized test runs of one
    fixed checkpoint per variant (test seeds 42, 123, and 2026),
    matching the main temporal evaluation.}
  \label{tab:daywise-results}
  \centering
  \scriptsize
  \setlength{\tabcolsep}{2.0pt}
  \renewcommand{\arraystretch}{1.08}
  \begin{tabular}{lcccccc}
    \toprule
    & \multicolumn{2}{c}{Energy $\downarrow$} & \multicolumn{2}{c}{Formal MMD $\downarrow$} & \multicolumn{2}{c}{DE@100 $\uparrow$} \\
    \cmidrule(lr){2-3}\cmidrule(lr){4-5}\cmidrule(lr){6-7}
    Method & Day~1 & Day~2 & Day~1 & Day~2 & Day~1 & Day~2 \\
    \midrule
    \textbf{TempoBridge Full Flow} & \textbf{0.131$\pm$0.006} & \textbf{0.128$\pm$0.004} & \textbf{0.373$\pm$0.015} & \textbf{0.353$\pm$0.014} & \textbf{0.501$\pm$0.013} & 0.624$\pm$0.009 \\
    Random-pairing Flow & 0.149$\pm$0.007 & 0.140$\pm$0.005 & 0.429$\pm$0.037 & 0.401$\pm$0.031 & 0.426$\pm$0.014 & \textbf{0.637$\pm$0.010} \\
    Static OT residual & 0.131$\pm$0.002 & 0.139$\pm$0.008 & 0.385$\pm$0.017 & 0.385$\pm$0.024 & 0.401$\pm$0.016 & 0.614$\pm$0.009 \\
    \bottomrule
  \end{tabular}
\end{table}

Table~\ref{tab:daywise-results} uses the same 2,009-gene space and
target-cardinality matching as the main temporal comparison. GENOT is not
included because its request-time evaluation uses source-matched output
cardinality. Full Flow leads Formal MMD and Energy at
both request times among the three reported transport variants. For DE@100,
Full Flow has the highest mean at Day~1, whereas Random-pairing Flow has the
highest mean at Day~2. The aggregate scores are the equal mean of the two
15-donor request-time strata within each randomized test run; Full Flow has
the highest aggregate DE@100 among these variants.

The known-transport synthetic experiment is separate from the shared-state
source-context construction in Appendix~\ref{app:controlled-source-context}.
It uses independently trained Full Flow and Static OT models with seeds
42, 123, and 2026, selected using endpoint data. Generation reads only Day~0
cells from four validation donors, with 512 cells per donor and requests at
Days~1 and 2. Model input and distribution scoring use the same
2,009-gene synthetic space. Scores are averaged over the eight donor/time
populations within each seed and then across seeds. Ground-truth
intermediate cells and correspondences are available only to the evaluator.
This diagnostic uses half the Energy V-statistic (unscaled expression) and
a single-bandwidth RBF MMD$^2$ V-statistic, with median pooled squared
distance as the squared bandwidth. Path and coupling metrics use intrinsic
coordinates recovered by the fixed generator-loading pseudoinverse;
coupling top-1 is the fraction whose nearest true intrinsic state is the
known matching cell. These scores are not directly comparable with the
real-data main table. Full Flow had lower intermediate-distribution Energy
($0.158\pm0.016$ versus $0.189\pm0.024$), lower MMD$^2$
($0.012\pm0.002$ versus $0.014\pm0.002$), and higher coupling top-1
accuracy ($0.339\pm0.015$ versus $0.223\pm0.036$). Static OT residual had
lower path RMSE, velocity error, and branch error. The Full Flow training formulation yielded better distribution recovery than
the Static OT residual formulation in this experiment, without lowering every
pathwise error.

\paragraph{Integration-horizon intervention.}
\label{sec:temporal-sensitivity}
We evaluated horizon interventions on the fixed 2,009-gene panel for three
validation donors, using the seed-42 checkpoint selected at update 8,500.
All interventions used 32 Heun steps and identical source inputs, with
generated population sizes matched to the observed targets. Energy was
averaged equally over the six donor--day requests.
Keeping the Day~0--Day~3 endpoint condition fixed, the prescribed
Day~1/2 horizons gave Energy $0.261$. Swapping the horizons increased Energy to
$0.457$; applying the Day~1 or Day~2 horizon to both requests gave $0.361$ and
$0.358$, and permutation gave $0.385$. This intervention tests dependence on
the requested integration progress, not whether the condition encoder learned
biologically calibrated Day~1 and Day~2 time.

\paragraph{Solver sensitivity.}
\label{sec:robustness}
\begin{table}[!htbp]
  \caption{Heun solver sensitivity for a frozen seed-42 Full Flow checkpoint
  on three validation donors (six requests), scored on the fixed 2,009-gene
  panel with target-matched cardinality. Scores average the six requests.
  One Heun step uses two velocity-network evaluations. Timing is seconds
  per 1,000 generated cells.}
  \label{tab:cost-robustness}
  \centering
  \small
  \begin{tabular}{rccc}
    \toprule
    Steps / calls & Energy $\downarrow$ & Formal MMD $\downarrow$ & Time (s) $\downarrow$ \\
    \midrule
     4 /  8 & 0.259 & 0.405 & 0.087 \\
     8 / 16 & 0.259 & 0.405 & 0.126 \\
    16 / 32 & 0.259 & 0.405 & 0.211 \\
    32 / 64 & 0.259 & 0.405 & 0.363 \\
    \bottomrule
  \end{tabular}
\end{table}

On validation seed 42, 8-step and 32-step Heun integration produced similar
Energy (0.259 and 0.259) and Formal MMD (0.405 and 0.405). Their measured
inference times were 0.126 and 0.363 seconds per 1,000 generated cells.
Timing covers source encoding, transport, and decoding from Day~0 input,
with one warmup and three repeats; it excludes disk I/O, host--device
transfer, serialization, and metric computation. These diagnostics use
three validation donors; the main table uses 15 held-out donors. Because
Heun evaluates the velocity network twice per step, the 32-step setting
makes 64 velocity-network calls.

\subsection{Additional protocol scope}
\label{app:additional-results}

In the daywise comparison, Full Flow and Random-pairing Flow had larger
Energy distances at Day~1 than at Day~2, whereas Static OT residual showed
the opposite ordering. Main-table
scores use target-matched output cardinality rather than source-matched
cardinality.

\section{Controlled Source-Context Ambiguity}
\label{app:controlled-source-context}

\subsection{Controlled construction}
\label{app:p0-1-construction}

The experiment is confined to a two-dimensional latent space and uses identity
encoders and decoders. For each replicate, contexts A and B contain the exact
same 128 shared source coordinates sampled from
$\mathcal N((0,0),0.15^2I)$. Each also contains 128 context-specific source
points, centered at $(-2,0)$ for A and $(2,0)$ for B. The corresponding shared
target components are sampled independently around $(0,2)$ and $(0,-2)$,
whereas the context-specific target components remain centered at $(-2,0)$ and
$(2,0)$. Each source and target population has 256 points. Targets
are shuffled after generation, and component labels are retained only for
analysis. Both contexts receive the fixed descriptor $c=(0,0)$.

The generator uses 64 training, 16 validation, and 32 test replicates
per context. Within a replicate, the maximum absolute difference between the
A/B shared source supports is exactly zero.The target branches differ only in
the population associated with those shared states. Their ideal centers are
separated by four latent units.

\subsection{Training protocol}
\label{app:p0-1-training}

The experiment directly instantiates \texttt{OTCFMTransition} with latent and
condition dimensions of 2, hidden width 128, no dropout, LayerNorm, and no
BatchNorm. All trainable arms contain 18,050 parameters. Training uses the
linear flow path, an entropic Sinkhorn coupling with $\varepsilon=0.05$ and 50
iterations, 64 OT cells per population, and batches of 16 populations. The
objective is flow-matching MSE only. AdamW uses learning rate 0.002, weight
decay $10^{-5}$, and gradient clipping at 1.0.

The maximum training length is 2,500 steps. Validation is performed every 250
steps using mean full-population Energy over validation replicates. Early
stopping begins after 500 steps, uses patience 6, and requires an improvement
of $10^{-4}$. Test predictions use a 32-step Heun solver with no initial noise.
Each trained arm contributes one validation-selected checkpoint, which is
held fixed across three test runs. No retraining or weight updates occur
between test runs. Hyperparameters were fixed before training, and no
post-result tuning was performed.

\subsection{Experimental arms}
\label{app:p0-1-arms}

The descriptor-only arm retains the same model and source particles as
TempoBridge but sets the pooled source-context slot to zero. TempoBridge uses
the mean of all 256 empirical source particles; the 64-point OT subsample
does not redefine this context. Swapped context uses the frozen
TempoBridge checkpoint and exchanges A/B empirical context vectors only at
inference, without retraining. The oracle arm is trained with fixed vectors
$(-1,0)$ and $(1,0)$ for contexts A and B; it provides a diagnostic reference with explicit context labels.

\subsection{Evaluation metrics and test-run aggregation}
\label{app:p0-1-results}

For each test replicate, $D_{\mathrm{ctx}}$ is the mean Euclidean distance
between the two predicted endpoints of each identical shared source point
under contexts A and B. Shared error is the mean of the distances from the
predicted shared-component centroids to the ideal centers $(0,2)$ and $(0,-2)$.
Branch accuracy counts a shared prediction as correct when its second
coordinate is strictly positive for A or strictly negative for B, then
averages over shared points and contexts; zero is counted as incorrect.
Energy evaluates the full predicted and target populations.

Table~\ref{tab:p0-1-main} reports the mean and sample SD across three test runs
of one fixed checkpoint per trained arm. Each run contributes one aggregate
score per metric. The sample SD uses the three run-level scores and
denominator $3-1$; it describes test-run variability conditional on the
selected checkpoint. The swapped-context setting uses the same TempoBridge
checkpoint as the correct-context setting, changing only the context input.

\begin{table}[!htbp]
    \centering
    \caption{Checkpoint-level results for the controlled source-context experiment. Seed indexes five training checkpoints, separately from the fixed-model randomized test repeats in Table~\ref{tab:p0-1-main}. Branch accuracies are tabulated as fractions. $S_{\mathrm{ctx}}$ is a post hoc directional metric computed from fixed predictions without retraining or checkpoint reselection.}
    \label{app:tab:p0-1-per-seed}
    \scriptsize
    \setlength{\tabcolsep}{2.8pt}
    \renewcommand{\arraystretch}{1.05}
    \begin{tabular}{@{}llrrrrrr@{}}
        \toprule
        Arm & Seed & $D_{\mathrm{ctx}}$ & $S_{\mathrm{ctx}}$ &
        Shared error & Branch acc. & Energy & MMD \\
        \midrule
        Descriptor-only & 42    & 0.000 & 0.000 & 2.001 & 0.500 & 0.457 & 0.273 \\
                        & 123   & 0.000 & 0.000 & 2.001 & 0.500 & 0.456 & 0.273 \\
                        & 2026  & 0.000 & 0.000 & 2.001 & 0.500 & 0.459 & 0.272 \\
                        & 31415 & 0.000 & 0.000 & 2.001 & 0.500 & 0.458 & 0.272 \\
                        & 27182 & 0.000 & 0.000 & 2.000 & 0.500 & 0.457 & 0.272 \\
        \addlinespace
        Swapped context & 42    & 3.899 & -3.882 & 3.941 & 0.013 & 3.430 & 0.779 \\
                        & 123   & 3.867 & -3.825 & 3.913 & 0.023 & 3.080 & 0.708 \\
                        & 2026  & 3.883 & -3.849 & 3.925 & 0.020 & 3.215 & 0.754 \\
                        & 31415 & 3.797 & -3.754 & 3.878 & 0.023 & 3.504 & 0.775 \\
                        & 27182 & 3.896 & -3.869 & 3.935 & 0.019 & 3.419 & 0.778 \\
        \addlinespace
        TempoBridge     & 42    & 3.899 & 3.882 & 0.079 & 0.987 & 0.006 & 0.000 \\
                        & 123   & 3.867 & 3.825 & 0.120 & 0.977 & 0.005 & 0.000 \\
                        & 2026  & 3.883 & 3.849 & 0.092 & 0.980 & 0.006 & 0.000 \\
                        & 31415 & 3.797 & 3.754 & 0.144 & 0.977 & 0.005 & 0.000 \\
                        & 27182 & 3.896 & 3.869 & 0.091 & 0.981 & 0.006 & 0.000 \\
        \addlinespace
        Oracle context  & 42    & 3.901 & 3.884 & 0.079 & 0.987 & 0.006 & 0.000 \\
                        & 123   & 3.883 & 3.840 & 0.118 & 0.975 & 0.006 & 0.000 \\
                        & 2026  & 3.875 & 3.841 & 0.096 & 0.979 & 0.006 & 0.000 \\
                        & 31415 & 3.929 & 3.894 & 0.100 & 0.987 & 0.005 & 0.000 \\
                        & 27182 & 3.893 & 3.865 & 0.092 & 0.981 & 0.006 & 0.000 \\
        \bottomrule
    \end{tabular}
\end{table}

\subsection{Velocity fields and model trajectories}
\label{app:p0-1-diagnostics}

\begin{figure}[!htbp]
    \centering
    \includegraphics[width=\linewidth]{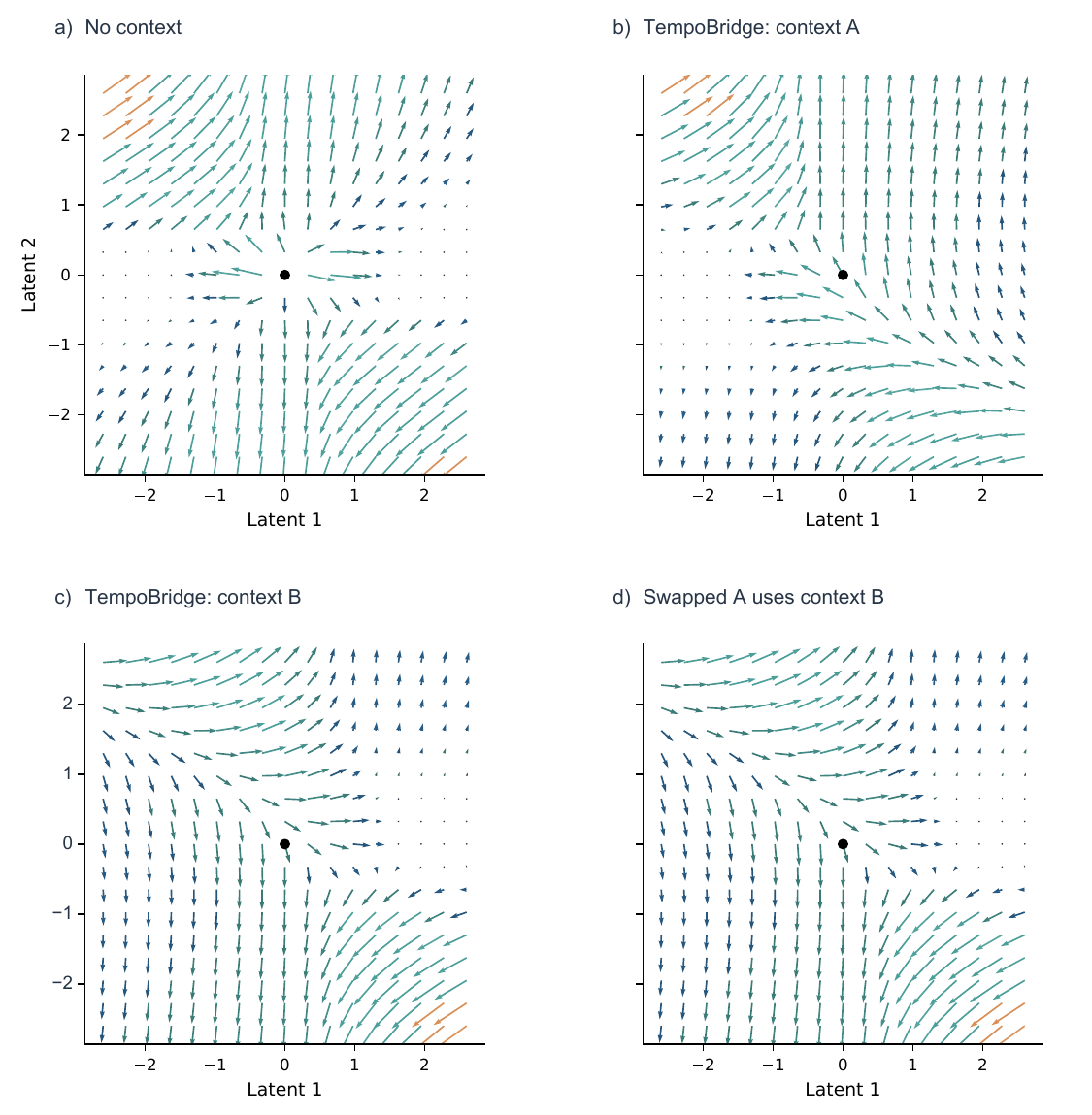}
    \caption{Velocity fields at $\tau=0.5$ for seed 42. The context-free
    field is identical for A and B at a shared state; empirical contexts produce
    distinct fields, and exchanging the context input exchanges the field.
    These are diagnostics in the controlled synthetic latent system.}
    \label{app:fig:p0-1-vector-fields}
\end{figure}

\begin{figure}[!htbp]
    \centering
    \includegraphics[width=0.92\linewidth]{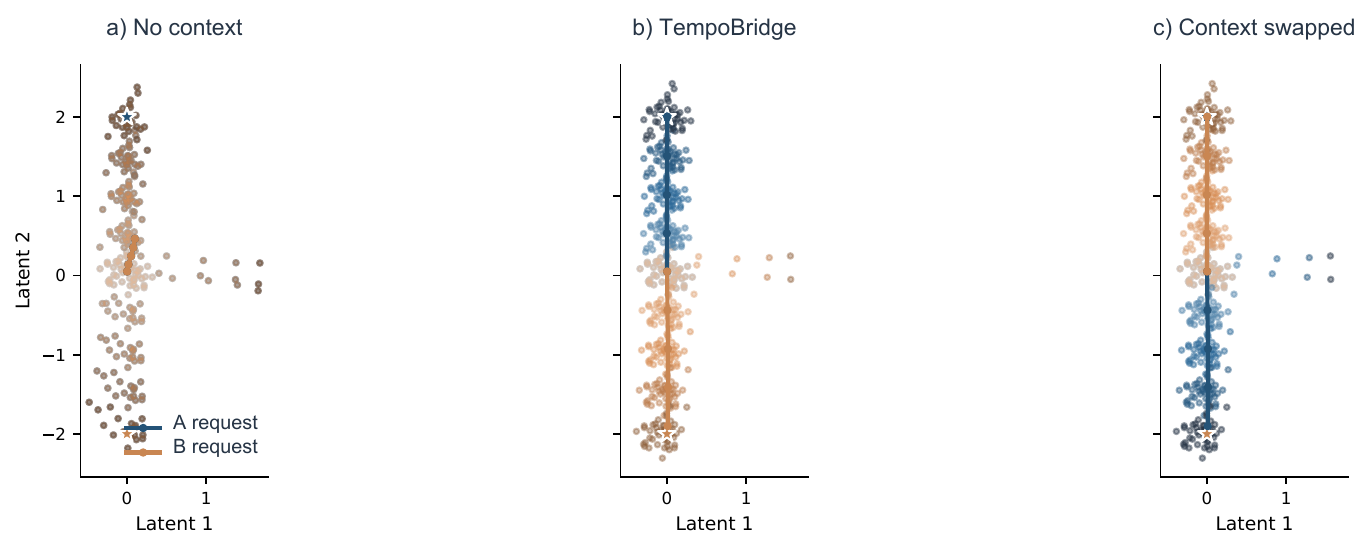}
    \caption{Model trajectories of the identical shared source states for seed
    42. TempoBridge separates the requested branches, while the context-free
    model overlaps and swapped context reverses the separation. The curves are
    synthetic ODE trajectories, not reconstructed biological lineages.}
    \label{app:fig:p0-1-trajectories}
    \label{fig:source_ambiguity}
\end{figure}

Figures~\ref{app:fig:p0-1-vector-fields} and~\ref{app:fig:p0-1-trajectories}
visualize the conditional fields and shared-state trajectories in the
controlled synthetic system.

\subsection{Velocity probe}
\label{app:p0-1-velocity-probe}

\begin{figure}[!htbp]
    \centering
    \includegraphics[width=\linewidth]{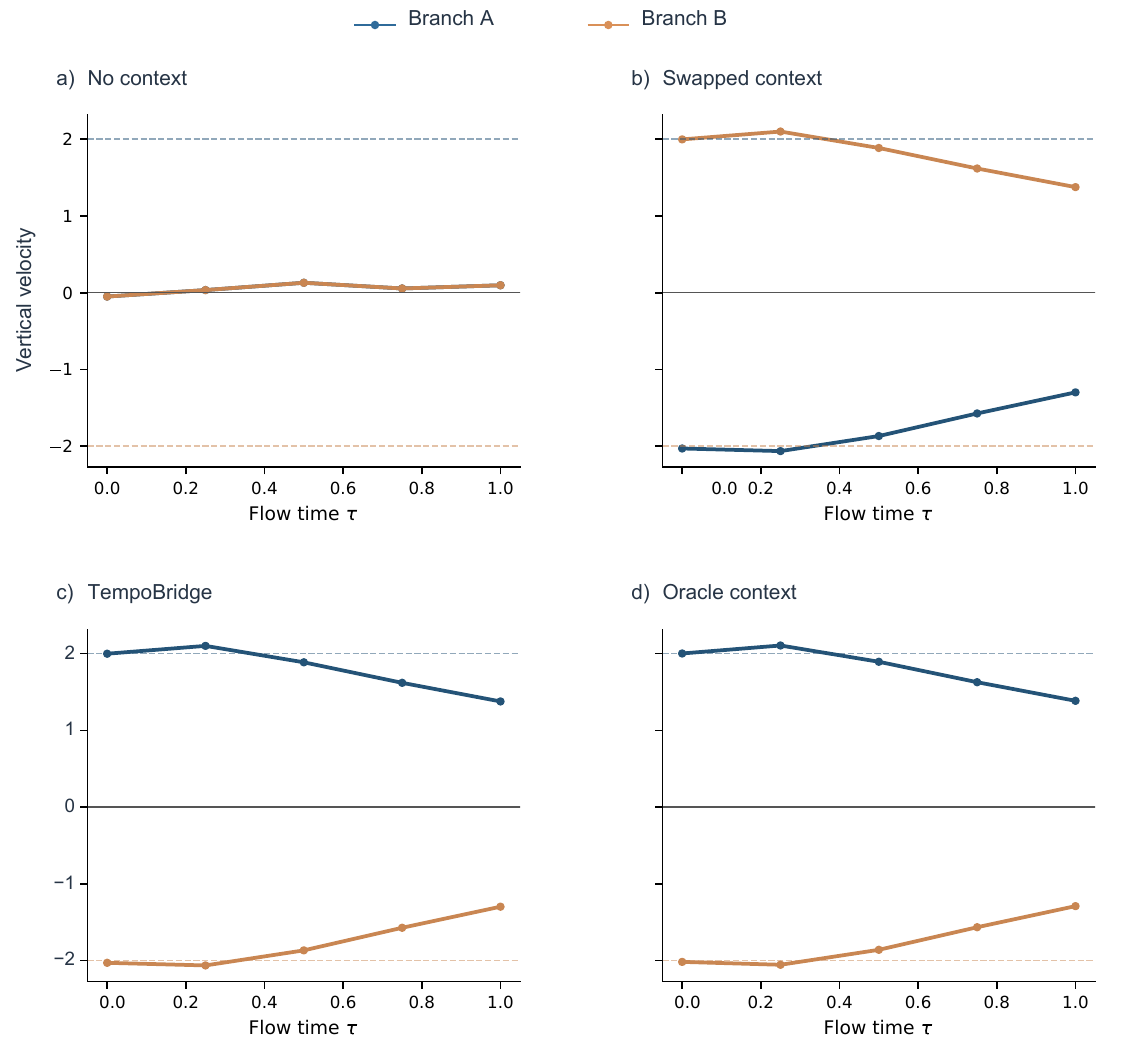}
    \caption{Velocity $y$-component at the identical state $z^\star=(0,0)$
    across flow time for seed 42. The context-free field returns identical
    velocities for A and B. TempoBridge assigns opposite directions, and
    swapping the empirical contexts reverses them. Dashed lines mark reference shared-component centroid displacements
    $+2$ and $-2$ over the unit transport interval, not exact conditional
    velocities at every probed time.}
    \label{app:fig:p0-1-velocity-probe}
\end{figure}

At $\tau=0$, the context-free velocity has $y=-0.050$ for both branches.
TempoBridge gives $1.999$ for A and $-2.030$ for B; the swapped-context
intervention exchanges these values. This fixed-state probe complements the
population-level metrics by checking the velocity interface directly.

\subsection{Controls for the ambiguity construction}
\label{app:p0-1-audit}

The A/B shared-support difference, condition difference, no-context
pointwise output difference, and maximum no-context
$D_{\mathrm{ctx}}$ are all zero. Parameter counts match across the
descriptor-only, TempoBridge, and oracle trainable arms, and no arm contains
BatchNorm. The comparisons in Table~\ref{tab:p0-1-main} assess shared-cell
branch accuracy and full-population Energy with no context, correct context,
and swapped context. All three test runs use one fixed checkpoint per trained
arm. Swapping context changes neither the TempoBridge parameters nor the
shared source coordinates.

\end{document}